\documentclass[conference]{IEEEtran}
\IEEEoverridecommandlockouts

\usepackage{hyperref}
\usepackage{xr-hyper}
\usepackage{%
  ./etc/kprema_newcommands}

\usepackage{cite}
\usepackage{url}
\usepackage{amsmath,amssymb,amsfonts, mathbbol}
\usepackage{algorithmic}
\usepackage{graphicx}
\usepackage{textcomp}
\usepackage{xcolor}
\usepackage{booktabs}
\def\BibTeX{{\rm B\kern-.05em{\sc i\kern-.025em b}\kern-.08em
    T\kern-.1667em\lower.7ex\hbox{E}\kern-.125emX}}

\begin{document}


\title{%
    Rarity of Singletons Matters: Unseen-Species and Missing-Mass Estimation in Small-Sample Regimes \\
    \thanks{%
        This work was supported by the French/US joint project LUCAS between the Agence Nationale de la Recherche (ANR) (grant \#ANR-25-CE23-2189) and the U.S. National Science Foundation (NSF) (grant \#s 2530255 and 2530256).}
}

\author{%
    \IEEEauthorblockN{%
        Pragatheeswaran Vipulanandan}
    \IEEEauthorblockA{%
        \textit{Dept. of Elect. and Comp. Engineering} \\
        \textit{University of Miami} \\
        Coral Gables, FL 33146, USA \\
        pxv245@miami.edu}
    \and
    \IEEEauthorblockN{%
        Kamal Premaratne}
    \IEEEauthorblockA{%
        \textit{Dept. of Elect. and Comp. Engineering} \\
        \textit{University of Miami} \\
        Coral Gables, FL 33146, USA \\
        kamal@miami.edu}
    \and
    \IEEEauthorblockN{%
        Dilip Sarkar}
    \IEEEauthorblockA{%
        \textit{Dept. of Comp. Science} \\
        \textit{University of Miami} \\
        Coral Gables, FL 33146, USA \\
        sarkar@miami.edu}
}

\maketitle


\begin{abstract}
Estimating missing-mass and unseen-species counts from finite samples relies critically on rare observations, particularly singletons. Yet in small-sample regime, these estimators can be highly \emph{fragile} in the sense that they may undergo large changes with even modest sampling fluctuations. We introduce a framework that explicitly accounts for this fragility and uses it to adaptively regularize standard missing-mass and unseen-species estimators. The resulting fragility-aware formulation provides a principled correction that becomes stronger precisely when the underlying rare-event statistics are least reliable. The entropy-KL optimization is strictly concave, admits a closed-form solution, requires no Poissonization, and can be applied as a lightweight correction to classical estimators such as Good-Turing and Good-Toulmin. Across 360 experiments spanning synthetic distributions and language corpora, our fragility-aware estimators remain competitive in stable regimes while consistently improving stability in long-tailed, singleton-dominated, small-sample regimes, with the largest gains occurring precisely where classical estimators are most fragile.
\end{abstract}



\section{Introduction}
\label{sec:intro}


A central question in statistical learning is how well a finite sample represents an underlying discrete distribution. information about the unobserved plays a critical role in addressing this question. In particular, the \emph{unseen-species} count (i.e., the number of species yet  undiscovered) and the \emph{missing-mass} (i.e., the probability mass associated with unseen-species) characterize sample coverage, generalization risk, and the likelihood of encountering new outcomes; in turn, they inform decisions regarding whether available data are sufficient or whether further sampling is warranted \cite{chao1984nonparametric, vipulanandan2026sensitivity}.


\subsection{Previous Work} 
\label{sec:prev_work}


Estimation of unseen-species and missing-mass dates to \cite{good1953population, good1956number}, with later work addressing development of estimators and performance guarantees under a variety of sampling models \cite{chao1984nonparametric, orlitsky2016optimal, hao2020optimal, kestemont2022forgotten}. Much of this work exploits occupancy statistics (particularly the frequencies of rare occurrences) to infer unseen-species and missing-mass. More recent results have looked at the information that these occupancy statistics carry and the fundamental limits of extrapolation from finite sample regimes \cite{orlitsky2016optimal, hao2020optimal, lee2025how}.

A practical issue however remains: in small or highly fragmented samples, the occupancy statistic of rare-events (i.e., singletons and doubletons) these estimators rely on can themselves be highly variable or \emph{fragile} in the sense that modest sampling fluctuations can produce substantial changes in missing-mass and unseen-species estimates \cite{weltz2024hidden}. This issue is particularly relevant in applications where only limited observations are available and sampling variability can dominate asymptotic behavior \cite{elena2025comprehensive}. Thus, beyond estimating the unseen-species and missing-mass, there is a need to quantify their \emph{fragility,} i.e., their sensitivity to sampling variability. 


\textbf{Contributions.}
We introduce a fragility-aware (FA) framework for missing-mass and unseen-species estimation. While only the main ideas and results are presented below, more detailed analysis, derivations, proofs, and additional experiments appear in  \cite{vipulanandan2027extended}. The principal contributions are:
\begin{enumerate}
    \item a singleton-based diagnostic that we call the \emph{fragility measure} to quantify when small-sample estimators are sensitive to limited or fragmented data or observations;  
    \item a concave \emph{entropy-KL correction} of a base estimator and closed-form estimators for both unseen-species count and their missing-mass; 
    \item \emph{theoretical analysis} to  characterize the expected fragility under multinomial sampling using a delta-method  approximation; and 
    \item extensive \emph{empirical validation} of our proposed framework across 360 experiments spanning synthetic distributions and language corpora, demonstrating improved stability in highly fragmented sampling regimes. 
\end{enumerate}


\section{Methodology}
\label{sec:methodology}

\begin{figure}[htpb]
    \centering
    \includegraphics[width = \columnwidth]{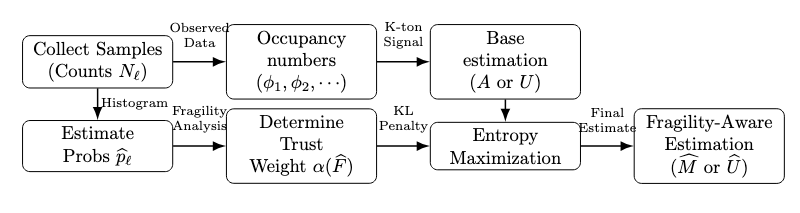}
    \caption{Overview of our fragility-aware estimation pipeline.}
    \label{fig:pipeline}
\end{figure}


\subsection{Preliminaries}
\label{sec:Preliminaries}


Take a set of species \(\mc{S} = \{s_{\ell}: 1 \leq \ell \leq S\}\) having the population frequencies \(\{p_{\ell}\}_{\ell=1}^S\). Given an $n$-length sample \(\uline{x}^{(n)} = \{x_1, \ldots, x_n\}\), the \emph{occupancy} of species $s_{\ell}$ (number of times species $s_{\ell}$ occurs) is
\begin{equation}
  N^{(n)}_{\ell}
    = \sum_{i = 1}^n 
      \mathbb{1}(X_i = s_{\ell}).
\end{equation}
The number of \emph{$k$-occupancies} (number of distinct species occurring exactly $k$ times) and the number of \emph{observed species} (distinct species observed at least once) are, respectively, 
\begin{equation}
  \label{eq:Phik}
  \Phi^{(n)}_k 
    = \sum_{s_{\ell} \in \mc{S}} 
      \mathbb{1}\!\bigl(N^{(n)}_{\ell} = k\bigr);\;\;
  \Phi_{\mr{obs}}^{(n)} 
    = \sum_{\ell} 
      \mathbb{1}(N_{\ell}^{(n)} > 0).
\end{equation}
Then, $\Phi_0^{(n)}$ is the number of missing species and $\Phi_1^{(n)}$ the number of singleton species. 

Henceforth, unless needed, we will drop the superscript $(n)$. 


\textbf{Missing-Mass.} 
\emph{Probability mass of $k$-occupancy} and \emph{probability mass of observed species} are \cite{good1953population}, respectively, 
\begin{equation}
  \label{eq:Mk}
  M_k
    \!=\!
      \sum_{s_{\ell} \in \mc{S}}
      p_{\ell}\, \mathbb{1}(N_{\ell} = k);\;\;
  M_{\mr{obs}} 
    \!=\! 
      \sum_{s_{\ell} \in \mc{S}}
      p_{\ell}\, \mathbb{1}(N_{\ell} > 0).
\end{equation}
Then, \(M_0\) is the \emph{missing-mass,} $M_1$ the \emph{singleton mass,} and \(1 - M_0\) the \emph{sample coverage} \cite{chao1984nonparametric}. As Good establishes \cite{good1953population}, $M_0$ is identical to the probability that the next generation expresses an unseen species.

The \emph{Good-Turin (GT) estimate} of \(M_k\) \cite{good1953population} is
\begin{equation}
  \label{eq:GTEstimate}
  \widehat{M}^{\mr{GT}}_{k}
    = (k + 1)/n
      \cdot
      \Phi_{k + 1}.
\end{equation}
Then, the GT estimate of missing-mass is $\wh{M}^{\mr{GT}}_0 = \Phi_1/n$. 


\textbf{Unseen-Species.} 
Unseen-species estimation refers to the number of new species that may appear in additional draws. For this purpose, Chao estimator and abundance-based coverage estimator (ACE) exploit singleton/doubleton ratios \cite{chao1984nonparametric, chao2000estimating} and Good-Toulmin estimator extrapolates expected discoveries \cite{good1956number}. Jackknife methods are used to reduce bias \cite{zahl1977jackknifing}. 

Suppose $U_{\mr{new}}$ denotes the number of yet unseen  new species that one sees in $n_{\mr{new}}$ additional sample draws (so that we end up with a sample of size $n + n_{\mr{new}}$). Then, the \emph{Good-Toulmin (GTo) estimate} of $U_{\mr{new}}$ \cite{good1956number} is
\begin{equation}
  \label{eq:GToEstimate}
  \wh{U}_{\mr{new}}^{(\mr{GTo})}
    = \sum_{k=1}^n 
      (-1)^{k+1}\, t^k\, \Phi_k,
      \tx{ where }
      n_{\mr{new}} = t\, n.
\end{equation}
This estimator of the number of new species $U_{\mr{new}}$ one may see in $n_{\mr{new}}$ additional samples is unbiased and performs well for $0 \leq t \leq 1$, but can become unstable for $t > 1$ \cite{good1956number}. 


\subsection{Motivation}


Existing missing-mass and unseen-species estimators depend strongly on low-frequency $k$-occupancies (i.e., low $k$ or rare species). As mentioned earlier, in small-sampling environments, these counts can be highly sensitive to sampling variability and hence fluctuate substantially. This gives rise to high fragility of the missing-mass and unseen-species estimates. For instance, the GT estimator \cite{good1953population} and its variants are singleton-driven and hence highly fragile at small $n$ values. So is the missing-mass estimator in \cite{lee2025how} because it is largely determined by the observed expected occupancy values. Poissonization is often used to simplify theoretical analysis to yield tractable error bounds \cite{mcallester2000convergence, drukh2005concentration, hao2020optimal}, but it ignores any multinomial dependence in the original sampling problem. Moreover, the bounds extracted  apply under asymptotic conditions \cite{hao2020optimal}. We therefore seek estimators that explicitly account for fragility in small-sample regimes with no recourse to Poissonization in order to retain the multinomial setting.

\begin{figure*}[t]
    \centering

    \includegraphics[width=0.80\textwidth]
    {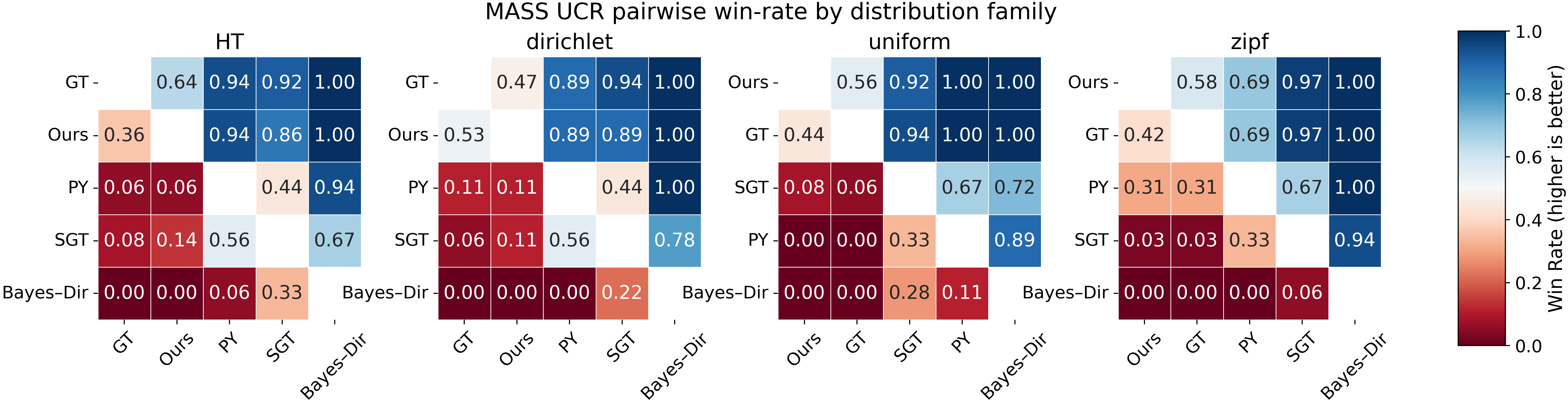}

    \par (a) Missing-mass estimation

    \vspace{0.2em}

    \includegraphics[width=0.80\textwidth]
    {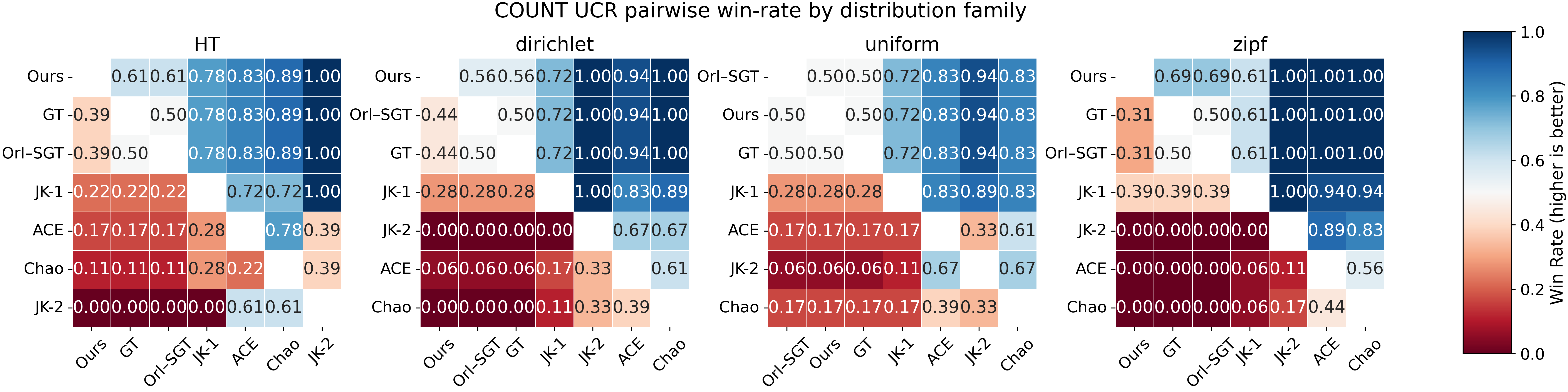}

    \par (b) Unseen-species prediction

 \caption{Per-family UCR win rates for missing-mass (top):
GT \cite{good1953population}, SGT \cite{gale1995good},
BD \cite{casella2002statistical}, PY \cite{pitman1997two};
and unseen-species (bottom): GTo \cite{good1956number},
Chao \cite{chao1984nonparametric}, ACE \cite{chao1992estimating},
Orl-SGT \cite{orlitsky2016optimal},
JK1/2 \cite{zahl1977jackknifing}.
Both include FA (ours). Entries indicate the fraction
of experiments where the row method achieves lower UCR
than the column method.}
    \label{fig:synth_family}
\end{figure*}

\subsection{Fragility-Aware Missing-Mass}
\label{subsec:fragility_mass}


Using the probability mass expressions $M_k$ and $M_{\mr{obs}}$ in \eqref{eq:Mk}, we define fragility and regularization strength as
\begin{equation}
  \label{eq:fragility_mass}
  F_{\mr{mass}}
    = M_1/M_{\mr{obs}};\;\;
  \alpha\,(F_{\mr{mass}})
    = \alpha_0\, e^{\gamma\, F_{\mr{mass}}},
\end{equation}
where $\alpha_{0} > 0$ and $\gamma < 0$.

Computation of fragility $F_{\mr{mass}} = M_1/M_{\mr{obs}}$ requires the population frequencies $\{p_{\ell}\}$, which we approximate via a Zipf-regularized empirical histogram \cite{baayen2001word}. We emphasize that this is solely to estimate $F_{\mr{mass}}$; the underlying GT and GTo estimators, and $F_{\mr{count}}$ introduced in \eqref{eq:fragility_count}, are all count-based.

The fragility-aware missing-mass estimate is computed via
\begin{equation}
  \label{eq:entropy_objective}
  \wh{M}_0^{\dag}
    = \arg\max_{M \in (0, 1)}
      \left\{
        h(M) - \alpha(F_{\mathrm{mass}})\, D_{\mathrm{KL}}(M \| \wh{M})
      \right\},
\end{equation}
with $\wh{M}$ a base estimator (we use the GT estimate  $\widehat{M}^{\mr{GT}}_0 = \Phi_1/n$ in \eqref{eq:GTEstimate}), $h$ the Bernoulli entropy, and $D_{\mathrm{KL}}(M \| \wh{M})$ the KL divergence  between Bernoulli parameters. 

The KL term preserves fidelity to $\wh{M}$. Note that, since $\gamma < 0$, increasing fragility reduces $\alpha(F_{\mr{mass}})$, thereby weakening the KL penalty and giving greater weight to the entropy term. In turn, as this entropy term increases, the estimate $\wh{M}_0^{\dag}$ moves toward the maximally noncommittal value $0.5$. In essence, under high fragility, the correction moves the estimate toward a less committed state instead of placing excessive confidence in an unstable base estimate.

Since the objective in \eqref{eq:entropy_objective} is differentiable and strictly concave in \(M\), its unique maximizer can be obtained by setting the derivative w.r.t. \(M\) to zero. The first-order condition yields the closed-form solution (see \cite{vipulanandan2027extended} for details) 
\begin{equation}
\label{eq:closed_form_mass}
  \mathrm{logit}(\wh{M}_0^{\dag})
    = \frac{\alpha(F_{\mr{mass}})}{1 + \alpha(F_{\mr{mass}})}\,
      \mathrm{logit}(\wh{M}),
\end{equation}
where $\mathrm{logit}(x)=\log\bigl(x/(1-x)\bigr)$. 



\subsection{Fragility-Aware Unseen-Species}
\label{subsec:fragility_count}


Whereas missing-mass quantifies the total probability mass assigned to $k$-occupancy species, unseen-species quantifies the number of distinct species that remain unobserved. To account for sampling fragility in this setting, consider the $k$-occupancy and total observed species counts $\Phi_k$ and $\Phi_{\mr{obs}}$ in \eqref{eq:Phik}. Then, analogous to \eqref{eq:fragility_mass}, we define count-based fragility and associated regularization strength as
\begin{equation}
  \label{eq:fragility_count}
  F_{\mr{count}} 
    = \Phi_1/\Phi_{\mr{obs}};\;\;
  \alpha(F_{\mr{count}})
    = \alpha_0\, e^{\gamma\, F_{\mr{count}}},
\end{equation}
with $\alpha_0 > 0$ and $\gamma < 0$.

The fragility-aware unseen-species count is then computed via \begin{equation}
  \label{eq:entropy_objective_count}
  \wh{u}_{\mr{new}}^{\dag}
    = \arg\max_{u \in (0, 1)} 
      \left\{
        h(u) 
          - \alpha\,(F_{\mr{count}})\, D_{\mathrm{KL}}(u \| \wh{u}_{\mr{new}})
      \right\},
\end{equation}
where, as we recall, $U_{\mr{new}}$ is the number of previously unobserved species discovered in $n_{\mr{new}}$ additional draws following an initial sample of size $n$; $u_{\mr{new}} = U_{\mr{new}}/U_{\max}$ is $U_{\mr{new}}$ normalized by $U_{\max}$, an upper bound for $U_{\mr{new}}$ so that $0 \leq U_{\mr{new}} \leq U_{\max}$ (e.g., one could use, as we do, $U_{\max} = n_{\mr{new}}$ for this). Similarly, $\wh{u}_{\mr{new}} = \wh{U}_{\mr{new}}/U_{\max}$ is the normalized value of $\wh{U}_{\mr{new}}$, a base estimator of $U_{\mr{new}}$ (we use the GTo estimator $\wh{U}_{\mr{new}}^{(\mr{GTo})}$ in \eqref{eq:GToEstimate} for this). 

As before, the KL term preserves fidelity to $\wh{u}_{\mr{new}}$. Since $\gamma < 0$, increasing fragility reduces $\alpha(F_{\mr{count}})$, thereby weakening the KL penalty and emphasizing the entropy term. As this entropy term increases, the estimate $\wh{u}_{\mr{new}}^{\dag}$ moves toward the maximally noncommittal value $0.5$; in turn, $\wh{U}_{\mr{new}}^{\dag} = U_{\max}\, \wh{u}_{\mr{new}}^{\dag}$, the corresponding fragility-aware estimate of the unseen species count, moves towards $0.5\, U_{\max}$. So, under high fragility, the correction moves the unseen-species estimate toward the least committed count, viz., the midpoint of the interval $[0,\, U_{\max}]$.

With an argument similar to that in Section~\ref{subsec:fragility_mass}, we can get the  closed-form solution (see supplementary \cite{vipulanandan2027extended} for details).
\begin{equation}
  \mathrm{logit}(\wh{u}_{\mr{new}}^{\dag}) 
    = \frac{\alpha(F_{\mr{count}})}{1 + \alpha(F_{\mr{count}})}\, \mathrm{logit}(\wh{u}_{\mr{new}}).
\end{equation}


\section{Experiments}
\label{sec:experiments}


We evaluate our estimates on synthetic distributions (with known $\{p_\ell\}$) and language corpora. Each trial draws an i.i.d.\ sample of size $n$, computes estimators from observed occupancies, then measures error on a disjoint future block of size $n_{\mr{new}}$ from the same population (Fig.~\ref{fig:pipeline}). Ground truth is empirical: $M_0$ is the fraction of holdout probability on species absent from the sample; $U_{\mr{new}}$ is the number of species in the holdout that were unseen in the sample.


\subsection{Synthetic Distributions}
\label{subsec:synthetic}


Synthetic populations span stable to fragmented regimes: 
    \textbf{Uniform} ($S \in \{200, 500, 1000\}$) as an equal-probability reference; 
    \textbf{Zipf} ($s \in \{0.5, 1.0, 1.5\}$) has varying tail heaviness; 
    \textbf{Dirichlet} ($\alpha \in \{0.3, 1.0, 3.0\}$) has random sparse/smooth populations; and
    \textbf{Head-tail mixtures} ($\rho \in \{0.7, 0.85, 0.95\}$) show dominant head plus diffused tail. 
    Sample sizes $n \in \{25, 50, 100, 200, 500, 1000\}$ cover small to moderate regimes.


\subsection{Real-World Datasets}
\label{subsec:real}


\textbf{Shakespeare} \cite{shakespeare_complete}: after lowercasing, punctuation normalization, and tokenization, uni-grams are species and corpus frequencies define the population; random length-$n$ subsamples simulate partial observation. \textbf{Google 1-gram} \cite{michel2011quantitative}: token corpus frequencies define the population with sharper tail sparsity than Shakespeare; subsamples at varying $n$ simulate incomplete coverage. Forecast horizons $r_{\mr{future}} = n_{\mr{new}}/n \in [0.03, 0.10]$ set holdout size and $U_{\max}=n_{\mr{new}}$; they define evaluation targets, not estimator inputs.


\subsection{Baselines}
\label{subsec:baselines}


\textbf{Missing-mass:} GT; smoothed GT (SGT); symmetric Dirichlet Bayes (BD); Pitman-Yor (PY) Bayes.

\textbf{Unseen counts:} Chao; ACE; GTo; Orl-SGT; first- and second-order jackknife (JK1, JK2).

Experiments\footnote{Code at  https://github.com/pragasv/Rarity-of-Singletons-Matters.git} ran on a workstation (64\,GB RAM, NVIDIA A6000). The fragility map $\alpha(F)=\alpha_0 e^{\gamma F}$ uses $\alpha_0>0$ as baseline trust in GT/GTo and $\gamma<0$ to weaken adjustment as fragility falls. We fix $(\alpha_0, \gamma) = (100, -1.0)$ globally across all configurations.


\subsection{Evaluation Metrics}
\label{subsec:metrics}


We report absolute error $|\widehat{M}_0 - M_0|$ and $|\widehat{U}_{\mr{new}} - U_{\mr{new}}|$ over 3000 trials per configuration. \textbf{Upper confidence risk (UCR)} bins fragility, with bin weights $w_i$, mean error $\mu_i$, and std.\ $\sigma_i$:
\begin{equation}
\label{eq:ucr}
  \mr{UCR}
    = \sum_i 
      w_i\, (\mu_i + z\, \sigma_i),
\end{equation}
where $z = 1.96$. Lower UCR indicates accuracy and stability. To capture effect size beyond win rates, we report normalized pairwise gain (baseline headroom gain):
\begin{equation}
\label{eq:gain}
  g_{ij}
    = \tanh\!\left(
        \gamma \cdot
        \frac{UCR_j - UCR_i}{|UCR_i| + |UCR_j|}
      \right),
\end{equation}
with $g_{ij}>0$ when row method $i$ lowers UCR versus column baseline $j$.


\section{Results}
\label{sec:results}


\subsection{Worked Example}
\label{sec:worked_example}



\begin{table*}[htpb]
    \caption{Worked example on the semantic-clustering query ($n = 15$, six clusters, $n_{\mr{new}} = 5$, $\{\Phi_1,\, \Phi_2,\, \Phi_{\mr{obs}}\} = \{3,\, 2,\, 6\}$, and $M_0 = 0.210$). The upper panel compares missing-mass estimators (GT, SGT, BD, PY, and FA), while the lower panel compares unseen-species count estimators (GTo, Chao, ACE, Orl-SGT, and FA). Our FA method achieves the lowest absolute error in both tasks.}
    \label{tab:worked_example}
    \centering
    \setlength{\tabcolsep}{3pt} \renewcommand{\arraystretch}{1.08}
    
    \begin{tabular}{l c cc ccccc}
        \toprule
        \textbf{Cluster}
            & \textbf{Count}
            & $p_s^{(r)}$ & $p_c^{(\ell)}$
            & $\widehat c^{(\mathrm{GT})}$ & $\widehat c^{(\mathrm{SGT})}$ & $\widehat c^{(\mathrm{BD})}$ & $\widehat c^{(\mathrm{PY})}$ & $\widehat c^{(\mathrm{FA})} (Ours)$ \\
        \midrule
        1. Russia
            & 1 
            & 0.1500 & 0.1500
            & 0.1519 & 0.1424 & 0.1582 & 0.1519 & 0.1512 \\
        2. S.~Arabia
            & 8 
            & 0.0275 & 0.2200
            & 0.2228 & 0.2089 & 0.2321 & 0.2228 & 0.2217 \\
        3. Iran
            & 2 
            & 0.0500 & 0.1000
            & 0.1013 & 0.0949 & 0.1055 & 0.1013 & 0.1008 \\
        4. Kuwait
            & 2 
            & 0.0250 & 0.0500
            & 0.0506 & 0.0475 & 0.0527 & 0.0506 & 0.0504 \\
        5. Qatar
            & 1 
            & 0.1400 & 0.1400
            & 0.1418 & 0.1329 & 0.1477 & 0.1418 & 0.1411 \\
        6. Iraq
            & 1 
            & 0.1300 & 0.1300
            & 0.1317 & 0.1234 & 0.1371 & 0.1317 & 0.1310 \\
        \midrule
        Unseen (tail)
            & -- 
            & -- & 0.2100
            & 0.2000 & 0.2500 & 0.1667 & 0.2000 & 0.2037 \\
        \textbf{Total}
            & 15 
            & -- & 1.0000
            & 1.0000 & 1.0000 & 1.0000 & 1.0000 & 1.0000 \\
        \midrule
        $\widehat M_0$
            & \multicolumn{3}{c}{$M_0 = 0.210$}
            & 0.2000 & 0.2500 & 0.1667 & 0.2000 & \textbf{0.2037} \\
        $|\widehat M_0-M_0|$
            & \multicolumn{2}{c}{}
            && 0.0100 & 0.0400 & 0.0433 & 0.0100 & \textbf{0.0063} \\
        \bottomrule
    \end{tabular}
    \vspace{0.6em}

    \begin{tabular}{l ccccc}
        \toprule
        \textbf{Unseen-count method}
            & \textbf{GTo} & \textbf{Chao} & \textbf{ACE} & \textbf{Orlitsky-SGT} & \textbf{FA} (Ours) \\
        \midrule
        $\widehat U_{\mr{new}}$ ($\mathbb{E}[U_{\mr{new}}] = 1.029$)
            & 1.000 & 2.250 & 5.786 & 0.833 & \textbf{1.018} \\
        \bottomrule
    \end{tabular}
\end{table*}

Table~\ref{tab:worked_example} shows the application of our proposed technique of missing-mass and unseen-count estimation to the same example in \cite{vipulanandan2026hallucination}. The input query \emph{``Which oil producer is a close ally of the United States?''} was presented, and repeated fifteen times, to an open-weight LLM. We choose an LLM setting because it embodies the small-sample, open-vocabulary regime that motivates this work: per-query generations are few, the number of possible
semantic clusters is unknown, and rare singletons can appear freely, with no fixed support size or tail assumptions.

The 15 responses form six semantic clusters ($\Phi_{\mr{obs}} = 6$) with occupancy counts
$N_{\ell}=\{1,8,2,2,1,1\}$ and occupancy statistics $\Phi_k = \{3,\, 2,\, \cdots, 1\}$ yielding three singletons ($\Phi_1$), two doubletons ($\Phi_2$) and one cluster observed eight times.
The observed clusters account for $\sum_{\ell} p_c^{(\ell)} = 0.79$ of the population
probability, leaving a true missing-mass of $M_0=0.21$. We model this remaining mass as 21 unseen meanings, each with probability $0.01$. For $n_{\mr{new}}=5$ additional
generations, the expected number of new semantics is $\mathbb{E}[U_{\mr{new}}]=21(1-0.99^5)=1.029$.

For each missing-mass estimator-GT, SGT, BD, PY, and our FA estimator, the estimated missing-mass $\widehat{M}_0$ yields adjusted \emph{observed coverage shares} $\widehat{c}^{(\ell)} = (1-\wh{M}_0)\, P_{\mr{obs}}^{(\ell)}/P_{\mr{obs}}$, with the residual $\wh{M}_0$ reserved for unseen meanings (\emph{Unseen} row in Table~\ref{tab:worked_example}). The lower panel in Table~\ref{tab:worked_example} separately compares unseen-species estimates from GTo, Chao, ACE, Orl-SGT, and our FA estimator. FA estimates use \eqref{eq:closed_form_mass}-\eqref{eq:entropy_objective_count} and attains the \emph{lowest absolute error of all methods} on both tasks. The richness estimators Chao/ACE target total support rather than the $n_{\mr{new}}$-horizon and overshoot. FA thus uses the extra fragility information ($p_c^{(\ell)}$ via $F_{\mr{mass}}$) to recalibrate GT when it is unreliable, the per-query face of the variance reduction quantified by our UCR analysis across repeated trials (see Sec.~\ref{sec:results}).


\subsection{Synthetic Distributions}
\label{subsec:synthetic_results}

\begin{figure}[t]
    \centering
    \begin{minipage}[b]{0.48\linewidth}
        \centering
        \includegraphics[width=\linewidth]
        {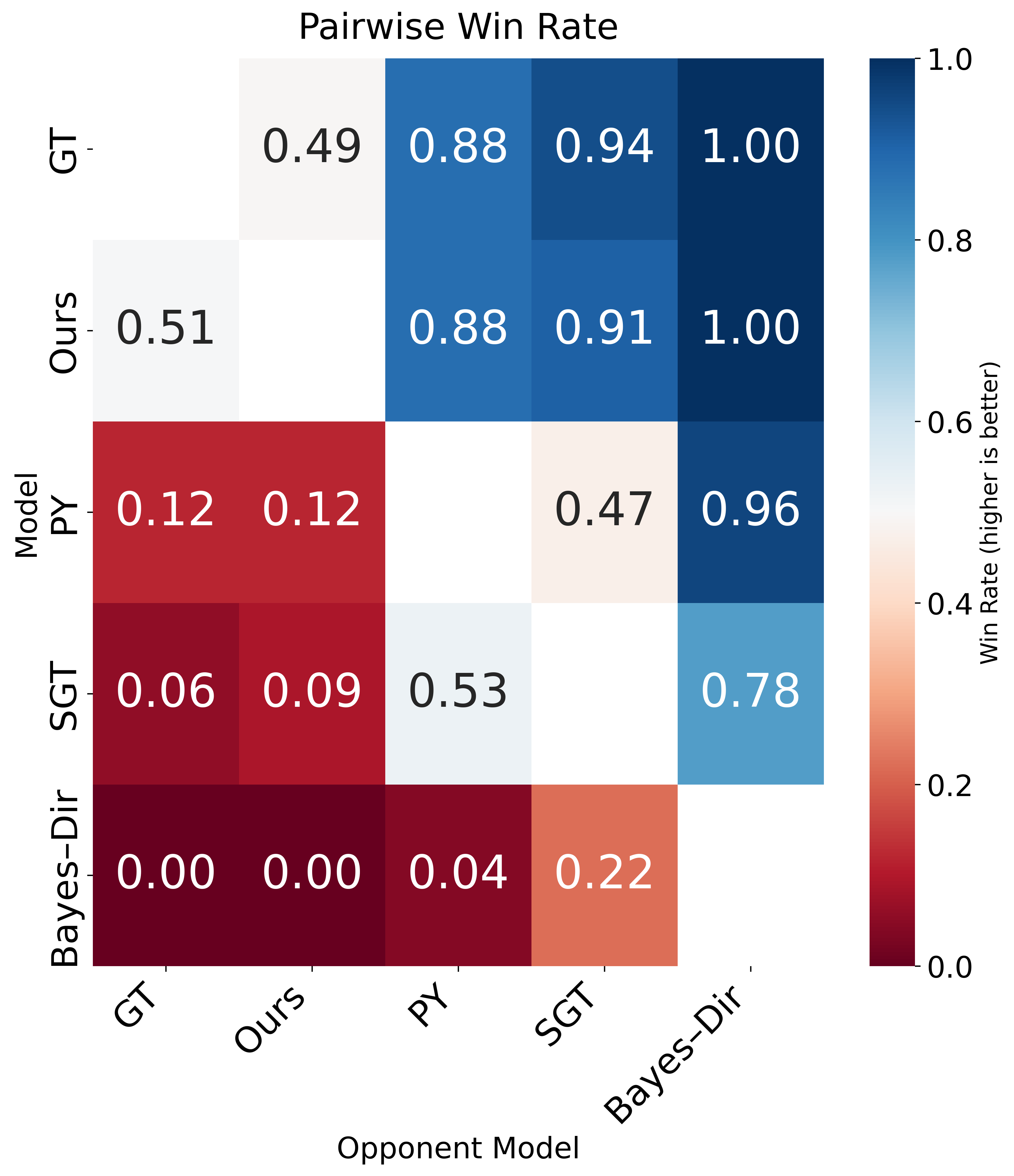}
        \par (a) Missing-mass win rate
    \end{minipage}
    \hfill
    \begin{minipage}[b]{0.48\linewidth}
        \centering
        \includegraphics[width=\linewidth]
        {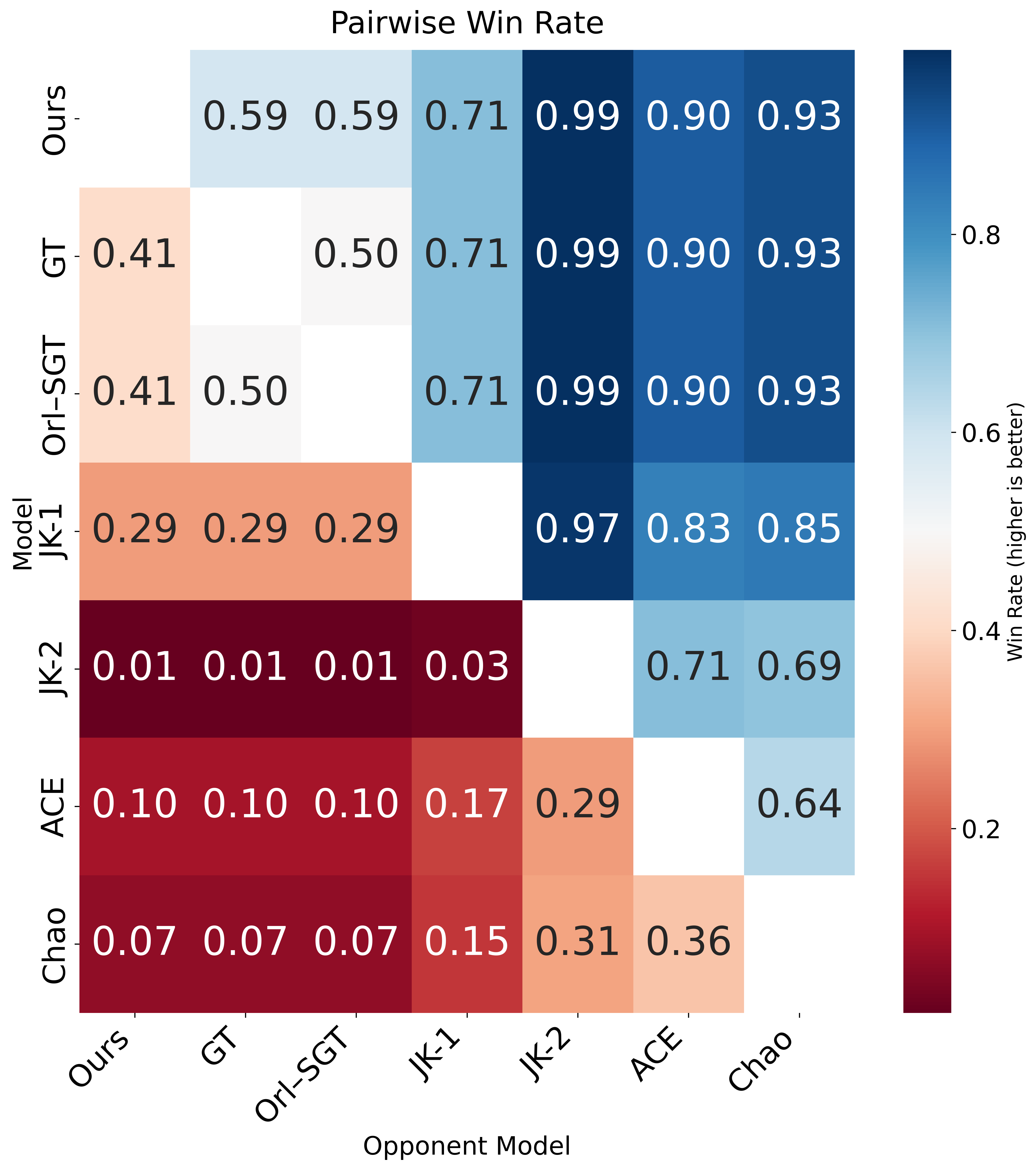}
        \par (b) Unseen-species win rate
    \end{minipage}
    \vspace{0.2em}
    \begin{minipage}[b]{0.48\linewidth}
        \centering
        \includegraphics[width=\linewidth]
        {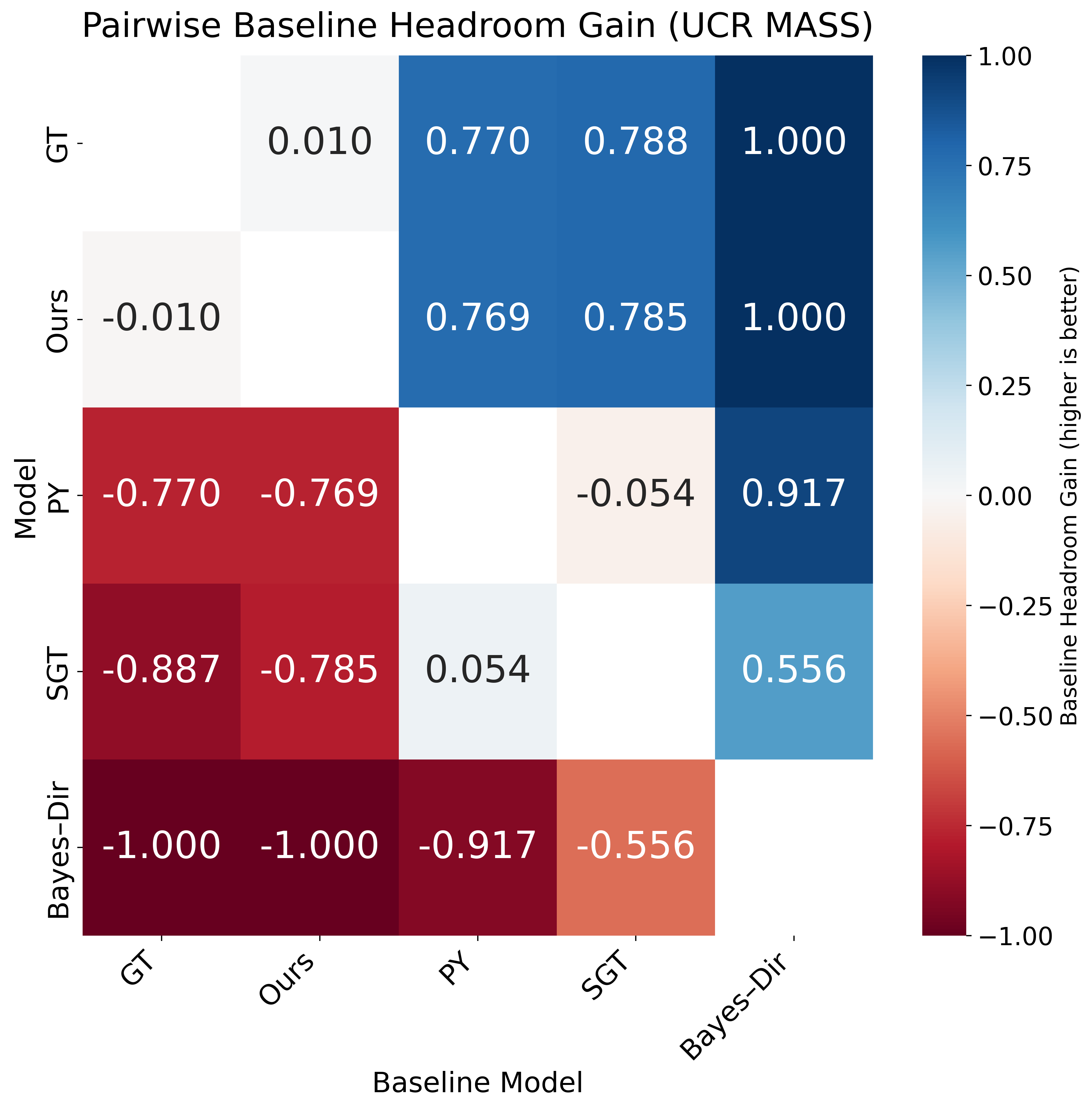}
        \par (c) Missing-mass gain
    \end{minipage}
    \hfill
    \begin{minipage}[b]{0.48\linewidth}
        \centering
        \includegraphics[width=\linewidth]
        {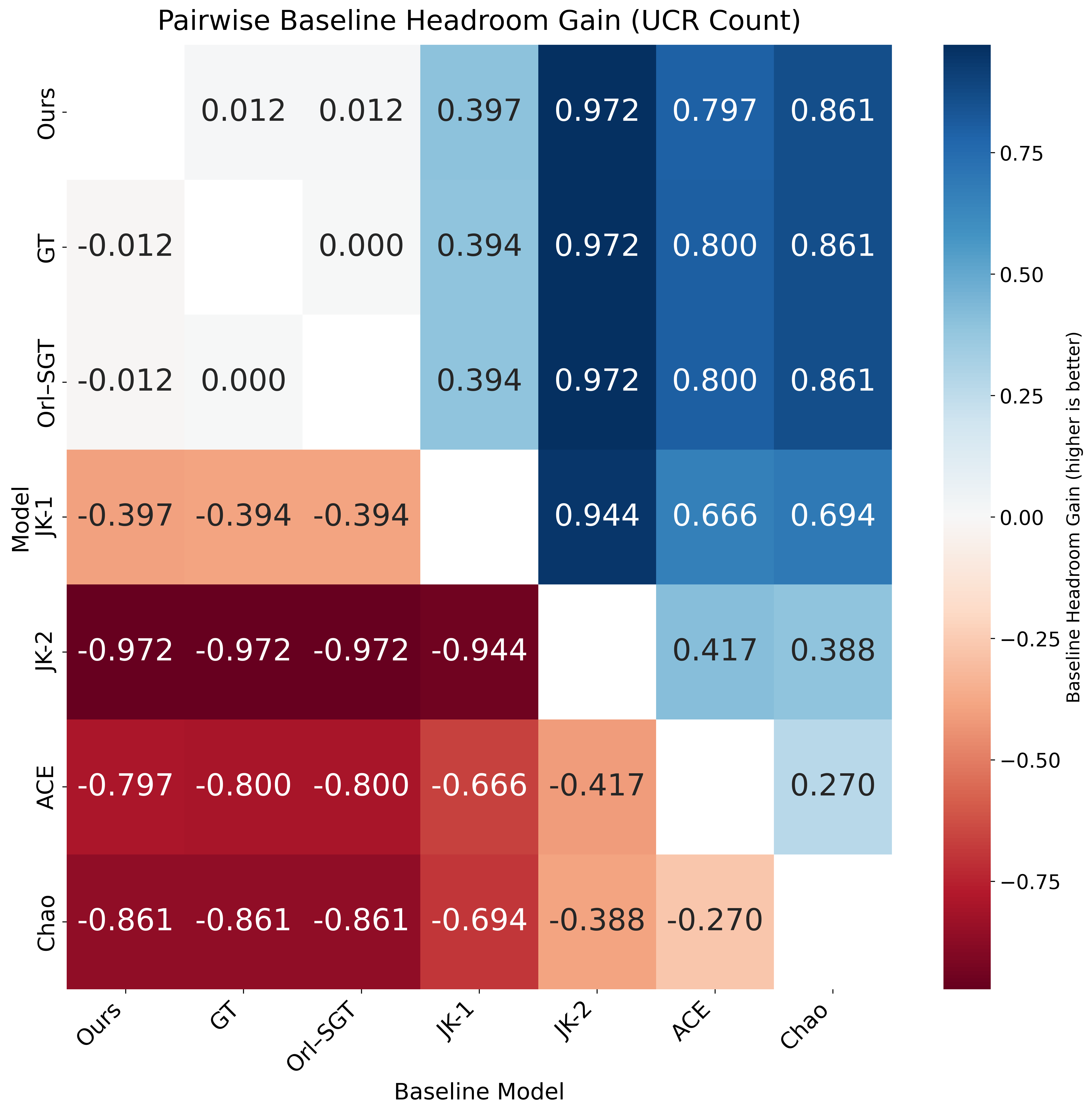}
        \par (d) Unseen-species gain
    \end{minipage}
    \caption{Synthetic overview (144 experiments). Top: pairwise UCR win rates; bottom: normalized pairwise gain $g_{ij}$ (see \eqref{eq:gain}). Blue indicates that the row method lowers UCR relative to the column baseline.}
    \label{fig:synth_gain_ovr}
\end{figure}

For the synthetic families of uniform, Zipf, Dirichlet, head-tail span smooth, skewed, and fragmented landscapes, Fig.~\ref{fig:synth_family} presents the per-family UCR win rates for missing-mass and unseen-species estimation, while Fig.~\ref{fig:synth_gain_ovr} summarizes the overall win rates and normalized gains $g_{ij}$.

\textbf{Missing-mass.} 
Win rates (Fig.~\ref{fig:synth_gain_ovr}a) and gain (Fig.~\ref{fig:synth_gain_ovr}c) agree: our estimator matches GT ($g \approx 0$) while substantially lowering UCR versus SGT, PY, and Bayes-Dir (Table~\ref{tab:synth_ucr_mass_zipf}). Per-family gains are largest on Zipf and Dirichlet; on head-tail mixtures, GT can edge ahead when singletons are stable (Supp.~\ref{app:HT_dist}).

\textbf{Unseen-counts.} 
Gains versus jackknife, ACE, and Chao are large ($g > 0.4$ in Fig.~\ref{fig:synth_gain_ovr}d), while differences among GT, Orlitsky-SGT, and our method are small. Win rates understate how poorly unstable baselines perform (Fig.~\ref{fig:synth_gain_ovr}b). Absolute UCR tables and head-tail analysis are in Supp.~\ref{App:extended_results_synth}.


\subsection{Real-World Datasets}
\label{subsec:realworld_results}


\begin{figure}[t]
    \centering
    \begin{minipage}[b]{0.48\linewidth}
        \centering
        \includegraphics[width=\linewidth]
        {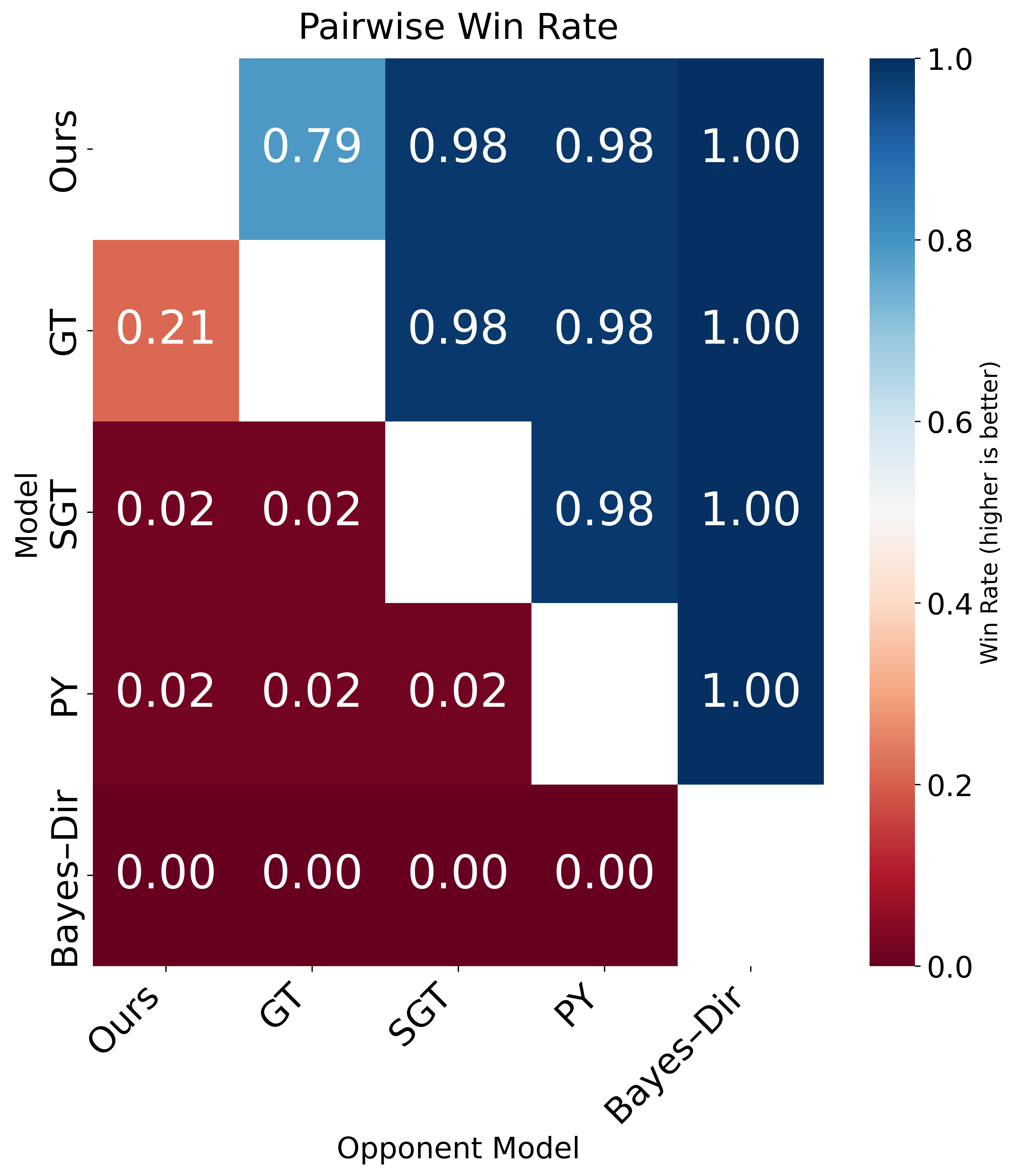}
        \par (a) Shakespeare: missing-mass
    \end{minipage}
    \hfill
    \begin{minipage}[b]{0.48\linewidth}
        \centering
        \includegraphics[width=\linewidth]
        {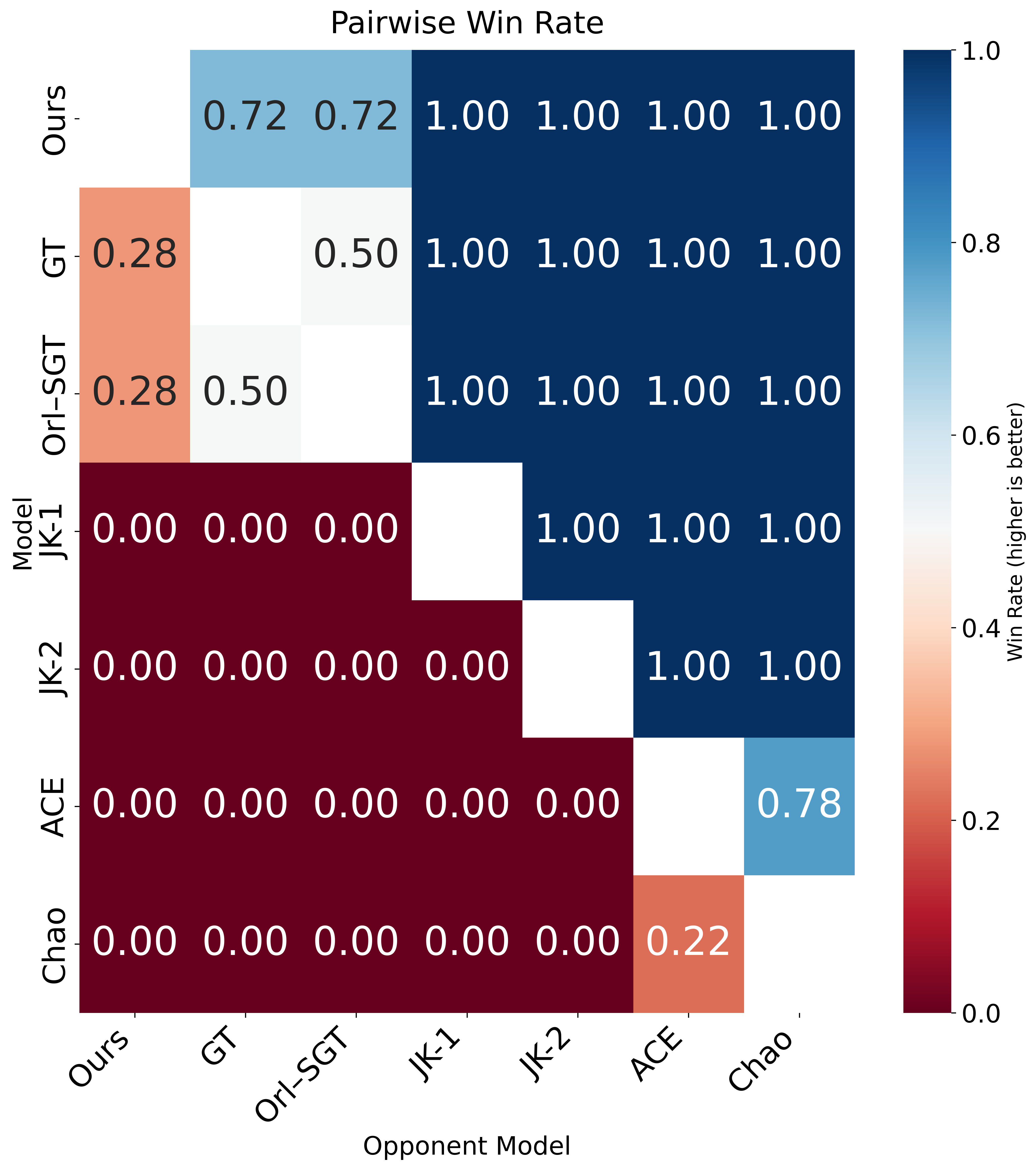}
        \par (b) Shakespeare: unseen-species
    \end{minipage}
    \label{fig:shake_ovr}
    \vspace{0.2em}
    \begin{minipage}[b]{0.48\linewidth}
        \centering
        \includegraphics[width=\linewidth]
        {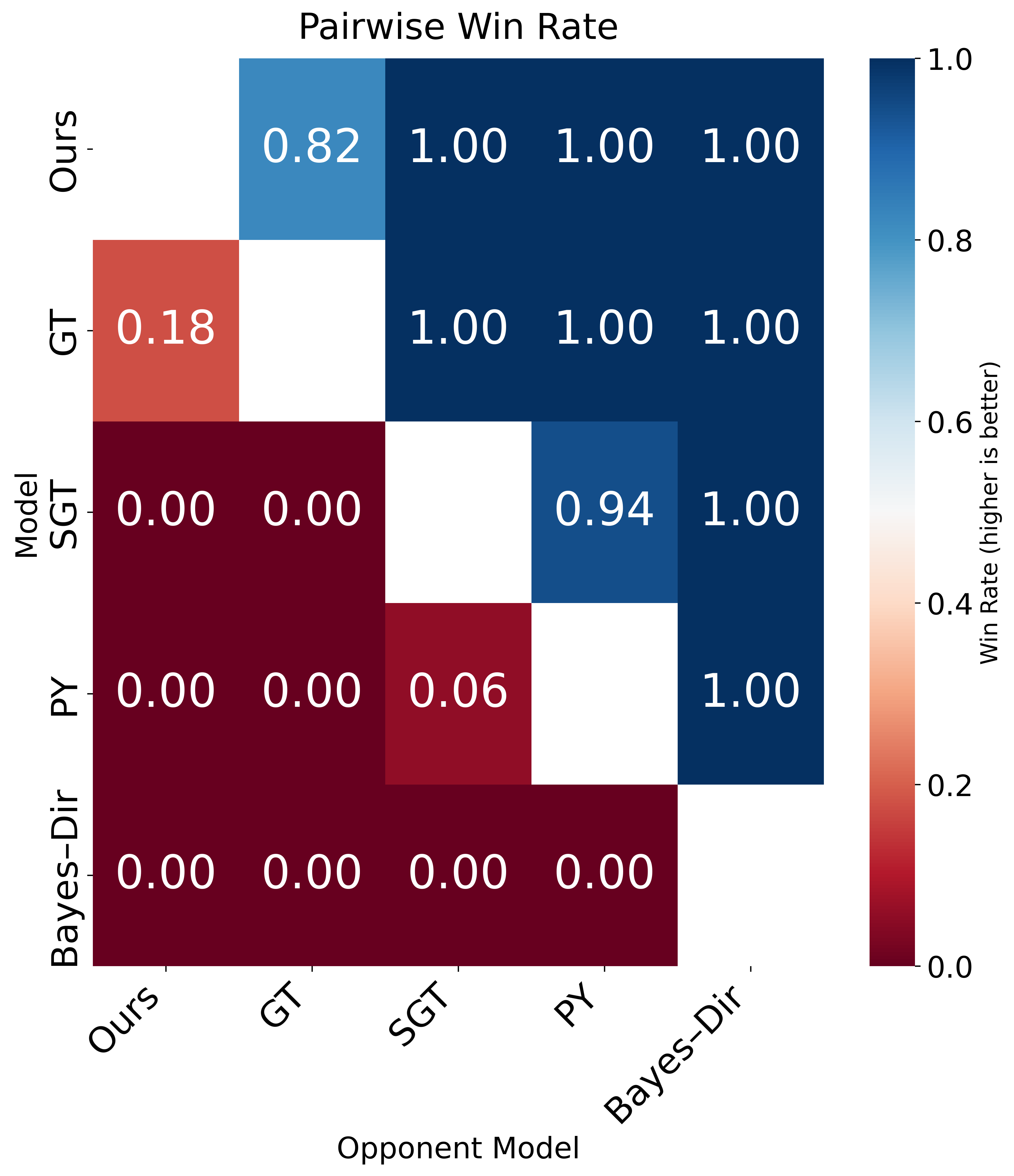}
        \par (c) Google: missing-mass
    \end{minipage}
    \hfill
    \begin{minipage}[b]{0.48\linewidth}
        \centering
        \includegraphics[width=\linewidth]
        {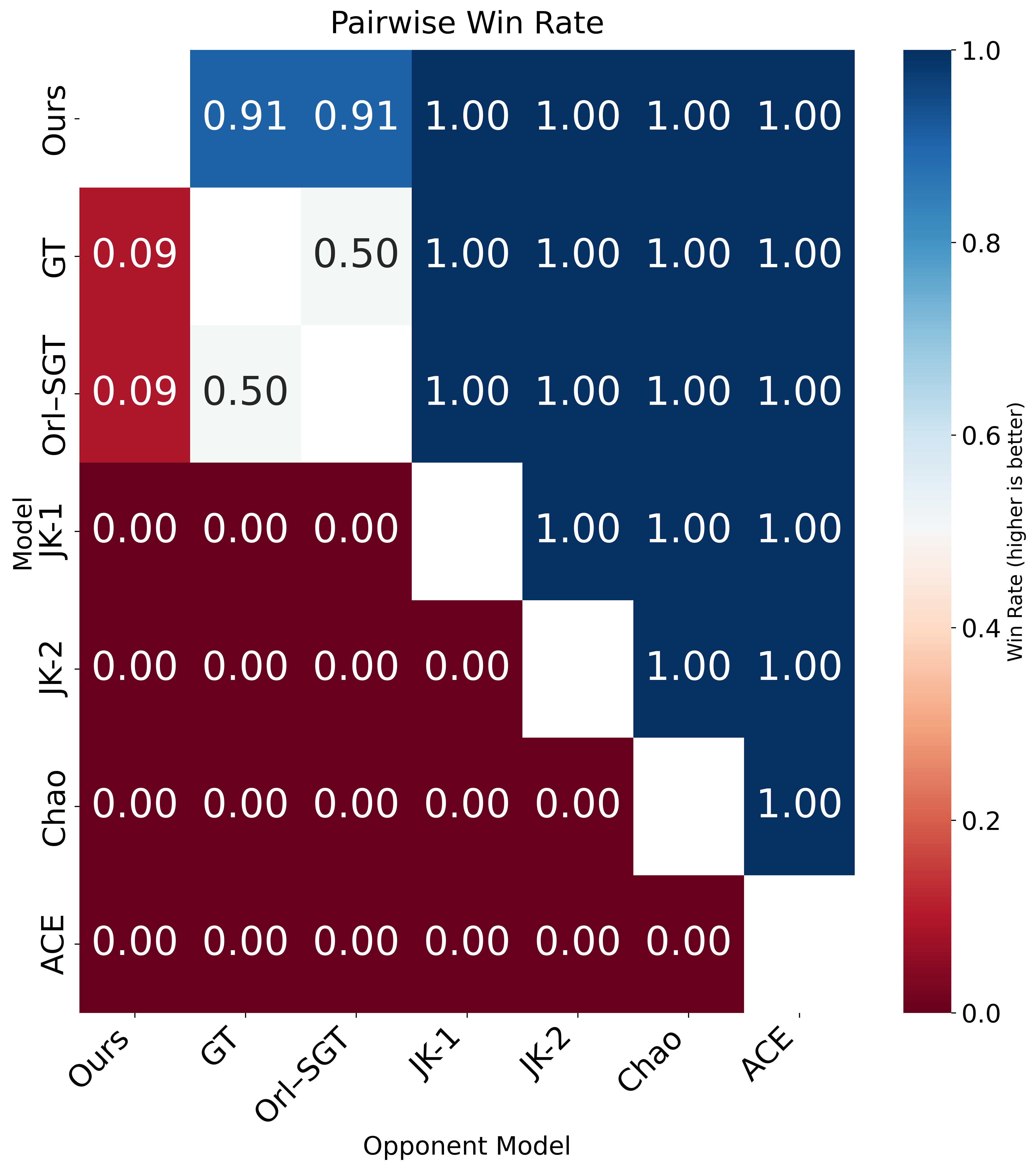}
        \par (d) Google: unseen-species
    \end{minipage}
    \label{fig:google_ovr}
    \caption{Pairwise UCR win rates across Shakespeare (top)
    and Google 1-gram (bottom), each with 108 experiments,
    pooled over $r_{\mathrm{future}}\in[0.03,0.10]$.
    Each entry $(i,j)$ denotes the fraction of experiments
    where row method $i$ outperforms column method $j$;
    values above $0.5$ indicate more frequent wins.
    Fragility-aware estimation performs best on unseen counts
    and the heavier-tailed Google corpus.}
    \label{fig:real_ovr}
\end{figure}

\textbf{Shakespeare Corpus (Project Gutenberg).} 
Fig.~\ref{fig:real_ovr}(a) shows competitiveness against classical methods and above GT on missing-mass. Fig.~\ref{fig:real_ovr}(b) showcases conservative adjustments beat aggressive extrapolation under long-tail noise. Additional details across forecast horizons are in Supp.~\ref{app:real_results_shake_mass}, Supp.~\ref{app:real_results_shake_count}.


\textbf{Google 1-Gram Corpus.}  
Google uni-grams have larger vocabulary and sharper head-tail separation. Fig.~\ref{fig:real_ovr}(c) showcases entropy-regularized adjustment outperforms GT-type estimators when frequent tokens dominate mass. Fig.~\ref{fig:real_ovr}(d) shows stability correction moderates over-discovery from aggressive smoothers. Additional details are across forecast horizons are provided in Supp.~\ref{app:real_results_google_mass} and Supp.~\ref{app:real_results_google_count}.


\section{Conclusion}


We introduce fragility-aware missing-mass and unseen-species
estimators that account for sampling instability through
singleton-based fragility measures and entropy-KL regularization.
The strictly concave objective admits a unique closed-form
solution without Poissonization, providing a lightweight
correction to classical estimators such as GT and GTo.
Across 360 experiments on synthetic distributions and language
corpora, our approach maintains competitive accuracy while
improving stability (lower UCR), particularly in long-tailed,
singleton-dominated, small-sample regimes.
These findings highlight the importance of rare-event
stability for reliable inference from incomplete data,
with applications to coverage estimation, biodiversity,
and uncertainty quantification in language models.

\clearpage


\section{AI Statement}


OpenAI's ChatGPT was used solely for language refinement and assistance with code development. The authors take full responsibility for the content of this work.

\raggedbottom
\bibliographystyle{IEEEtran}
\bibliography{IEEEabrv, references}


\end{document}


\title{Supplementary Materials: Rarity of Singletons Matters}

\author{Pragatheeswaran Vipulanandan,~\IEEEmembership{Member,~IEEE,} Kamal Premaratne,~\IEEEmembership{Senior Member,~IEEE} and Dilip Sarkar,~\IEEEmembership{Senior Member,~IEEE}%
\thanks{J. Doe and J. Smith are with the Department of Electrical Engineering, University of Technology, City, Country (e-mail: doe@uni.edu).}}

\maketitle
\beginsupplement 

This document provides supplementary details, extended derivations, and additional experimental evaluations to accompany the main manuscript.

\section{Properties of the Fragility Adjustment}
\label{App:fragility_properties}

Let $A\in(0,1)$ denote a base estimate and
$F\in[0,1]$ the corresponding fragility measure,
where $F=F_{\mr{mass}}$ for missing-mass and
$F=F_{\mr{count}}$ for unseen-species count estimation.

Both fragility-aware estimators maximize the
generic entropy--KL objective
\begin{equation}
J(q)=h(q)-\alpha(F)D_{\mathrm{KL}}(q\|A),
\qquad q\in(0,1),
\end{equation}
where $q$ represents either the missing-mass
parameter $M$ or the normalized unseen-species
parameter $u$.

The corresponding fragility-aware estimate
$\wh{q}^{\dag}$ is the unique maximizer
of $J(q)$. Thus, all structural properties derived below therefore apply to both the
fragility-aware missing-mass estimator and the fragility-aware unseen-species
estimator.

\subsection{Convexity of the objective}

\begin{lemma}
\label{lem:concavity}
Let $A\in(0,1)$ and $\alpha(F)>0$.
The objective
\[
J(q)
=
h(q)
-
\alpha(F)
\,
D_{\mathrm{KL}}(q\|A)
\]
is strictly concave in $q\in(0,1)$.
\end{lemma}

\begin{proof}
The Bernoulli entropy satisfies
\[
h(q)=-q\log q-(1-q)\log(1-q),
\]
with derivatives
\[
h'(q)=\log\!\frac{1-q}{q},
\qquad
h''(q)
=
-\frac{1}{q}-\frac{1}{1-q}.
\]

The KL divergence in its first argument is
\[
D_{\mathrm{KL}}(q\|A)
=
q\log\!\frac{q}{A}
+
(1-q)\log\!\frac{1-q}{1-A},
\]
whose derivatives are
\[
\frac{\partial}{\partial q}D_{\mathrm{KL}}
=
\log\!\frac{q(1-A)}{A(1-q)},
\qquad
\frac{\partial^2}{\partial q^2}D_{\mathrm{KL}}
=
\frac{1}{q}+\frac{1}{1-q}.
\]

Therefore the second derivative of $J$ is
\[
J''(q)
=
h''(q)
-
\alpha(F)
\frac{\partial^2}{\partial q^2}D_{\mathrm{KL}}
=
-\bigl(1+\alpha(F)\bigr)
\left(\frac{1}{q}+\frac{1}{1-q}\right).
\]

Since $\alpha(F)>0$ and
$\frac{1}{q}+\frac{1}{1-q}>0$ for all $q\in(0,1)$,
it follows that $J''(q)<0$ everywhere on $(0,1)$.
Hence $J$ is strictly concave and admits a unique maximizer.
\end{proof}

\subsection{Limit Analysis}

The following result characterizes the limiting behavior
of the generic fragility-aware estimator $\wh{q}^{\dag}$
introduced in Section~\ref{App:fragility_properties}.
Here, $q=M$ with $A=\wh{M}$ for missing-mass estimation,
and $q=u$ with $A=\wh{u}_{\mr{new}}$ for normalized
unseen-species estimation.

\textbf{Lemma.}
Let $A \in (0, 1)$ and let $\wh{q}^{\dag}(\alpha)$
denote the unique maximizer of the generic objective
$J(q)$ for a given $\alpha > 0$. Then
$$
\lim_{\alpha \to \infty}
\wh{q}^{\dag}(\alpha)
= A.
$$

\emph{Proof: Write the objective as
$$
J_\alpha(q)
= h(q) - \alpha\, D_{\mathrm{KL}}(q\|A),
$$
where
$$
h(q)
= -q \log\,(q) - (1 - q)\, \log\,(1 - q);\;\;
D_{\mathrm{KL}}(q \| A)
= q\, \log\,\frac{q}{A}
+ (1 - q)\, \log\frac{1 - q}{1 - A}.
$$
But $J_\alpha$ is strictly concave on $(0,1)$
(see Lemma~\ref{lem:concavity}).
So, the maximizer $\wh{q}^{\dag}(\alpha)$
is unique and satisfies the first-order condition
$J_\alpha'(\wh{q}^{\dag}(\alpha)) = 0$.
A direct calculation of the derivatives $h'(q)$
and $\partial D_{\mathrm{KL}}(q \| A)/\partial q$
yields that $\wh{q}^{\dag}(\alpha)$ satisfies
$$
\log\frac{1 - \wh{q}^{\dag}(\alpha)}
{\wh{q}^{\dag}(\alpha)}
= \alpha\, \log\frac{
\wh{q}^{\dag}(\alpha)\, (1 - A)}
{(1 - \wh{q}^{\dag}(\alpha))\, A}.
$$
We can express this as
$$
-\mathrm{logit}\,(\wh{q}^{\dag}(\alpha))
= \alpha\, (
\mathrm{logit}\,(\wh{q}^{\dag}(\alpha))
- \mathrm{logit}\,(A)),
$$
where $\mathrm{logit}(\centerdot)$ is the logit
function (e.g., $\mathrm{logit}\,(A)
= \log\,(A/(1 - A))$).
This gives us
$$
\mathrm{logit}\,(\wh{q}^{\dag}(\alpha))
= \frac{\alpha}{\alpha + 1}\,
\mathrm{logit}\,(A).
$$
So, as $\alpha \to \infty$,
$\mathrm{logit}\,(\wh{q}^{\dag}(\alpha))
\to \mathrm{logit}\,(A)$
and hence $\wh{q}^{\dag}(\alpha) \to A$.
\hfill Q.E.D.}

This result applies to both estimation tasks.
For missing-mass estimation, the fragility-aware
estimate $\wh{M}_0^{\dag}$ converges to its base
estimate $\wh{M}$ as $\alpha\to\infty$.
Similarly, for unseen-species estimation, the
normalized estimate $\wh{u}_{\mr{new}}^{\dag}$
converges to $\wh{u}_{\mr{new}}$, implying that
the corresponding unnormalized estimate
$\wh{U}_{\mr{new}}^{\dag}$ converges to
$\wh{U}_{\mr{new}}$.
Conversely, as $\alpha\to0^+$, the closed-form
solution shows that both normalized estimators
approach $1/2$, corresponding to a missing-mass
estimate of $1/2$ and an unseen-species count
estimate of $U_{\max}/2$, respectively.

\subsection{Expectation of Fragility}
\label{subsec:expected_fragility}

For missing-mass fragility $\widehat{F}_{\mr{mass}} = P_1/P_{\mr{obs}}$, the ratio is nonlinear in correlated sums and $\mathbb{E}[\widehat{F}_{\mr{mass}}]$ has no exact closed form. A second-order delta approximation \cite{Cochran1977Sampling} gives
\begin{align}
\label{eq:approxesti_F}
\mathbb{E}[\widehat{F}_{\mr{mass}}]
&\approx
\frac{\mu_A}{\mu_B}
-
\frac{\Cov(P_1,P_{\mr{obs}})}{\mu_B^2}
+
\frac{\mu_A\,\Var(P_{\mr{obs}})}{\mu_B^3},
\end{align}
with $\mu_A=\mathbb{E}[P_1]$, $\mu_B=\mathbb{E}[P_{\mr{obs}}]$. Under $n$ multinomial draws and $N_\ell\sim\mr{Binomial}(n,p_\ell)$,
\begin{align}
\label{eq:muAB_frag}
\mu_A
&= n\sum_{\ell=1}^S p_\ell^2(1-p_\ell)^{n-1},
&
\mu_B
&= 1-\sum_{\ell=1}^S p_\ell(1-p_\ell)^n.
\end{align}
Variances and covariances follow from binomial indicator moments (e.g., for $\ell\neq m$, $\mathbb{E}[\mathbb{1}(N_\ell>0)\mathbb{1}(N_m>0)] = 1-(1-p_\ell)^n-(1-p_m)^n+(1-p_\ell-p_m)^n$). The same structure applies to $\widehat{F}_{\mr{count}}=\Phi_1/\Phi_{\mr{obs}}$ with count-based indicators.

When many species carry small $p_\ell$ at fixed $n$, singleton mass dominates and $\mu_A/\mu_B$ rises---predicting high fragility and unstable GT/GTo corrections. As $n$ increases, $\mu_B\to 1$ and expected fragility typically falls unless the tail remains heavy (Zipf/Dirichlet). Structured head--tail splits can instead stabilize singleton statistics, explaining when GT remains competitive (App.~\ref{app:HT_dist}).

\section{Extended Synthetic Results}
\label{App:extended_results_synth}

\subsection{Zipf distributions}
\label{subsec:zipf_extended}

Table~\ref{tab:synth_ucr_mass_zipf} reports UCR for missing-mass estimation on Zipf populations ($p_i \propto i^{-s}$, $s \in \{0.5,1.0,1.5\}$) across sample sizes $N$. Ours matches or beats GT at most $(s,N)$ while substantially outperforming SGT and Bayesian baselines at small $n$; absolute gaps are modest versus GT but large versus PY/Bayes--Dir.

\begin{table}[htbp]
\centering
\scriptsize
\setlength{\tabcolsep}{3.5pt}
\renewcommand{\arraystretch}{1.05}
\begin{tabular}{llrrrrrr}
\toprule
family & parameter & $N$ & Ours & GT & SGT & BD & PY \\
\midrule
zipf & $s=0.5$ & 25 & \textbf{0.149} & \textbf{0.149} & \textbf{0.149} & 0.510 & 0.978 \\
zipf & $s=0.5$ & 50 & 0.133 & \textbf{0.132} & 0.201 & 0.463 & 0.914 \\
zipf & $s=0.5$ & 100 & \textbf{0.123} & \textbf{0.123} & 0.242 & 0.389 & 0.803 \\
zipf & $s=0.5$ & 200 & \textbf{0.099} & 0.100 & 0.194 & 0.259 & 0.628 \\
zipf & $s=0.5$ & 500 & 0.065 & \textbf{0.064} & 0.280 & 0.081 & 0.356 \\
zipf & $s=0.5$ & 1000 & \textbf{0.033} & \textbf{0.033} & 0.255 & 0.076 & 0.168 \\
zipf & $s=1.0$ & 25 & \textbf{0.271} & 0.272 & 0.395 & 0.406 & 0.758 \\
zipf & $s=1.0$ & 50 & \textbf{0.199} & 0.200 & 0.375 & 0.321 & 0.644 \\
zipf & $s=1.0$ & 100 & \textbf{0.121} & 0.122 & 0.259 & 0.221 & 0.504 \\
zipf & $s=1.0$ & 200 & \textbf{0.094} & 0.095 & 0.182 & 0.175 & 0.402 \\
zipf & $s=1.0$ & 500 & 0.050 & \textbf{0.049} & 0.068 & 0.069 & 0.240 \\
zipf & $s=1.0$ & 1000 & 0.032 & \textbf{0.031} & 0.080 & 0.028 & 0.153 \\
zipf & $s=1.5$ & 25 & 0.251 & 0.252 & 0.292 & \textbf{0.235} & 0.434 \\
zipf & $s=1.5$ & 50 & 0.147 & 0.147 & 0.181 & \textbf{0.138} & 0.313 \\
zipf & $s=1.5$ & 100 & 0.109 & 0.109 & 0.139 & \textbf{0.100} & 0.234 \\
zipf & $s=1.5$ & 200 & \textbf{0.062} & \textbf{0.062} & 0.086 & \textbf{0.062} & 0.161 \\
zipf & $s=1.5$ & 500 & \textbf{0.034} & \textbf{0.034} & 0.049 & 0.035 & 0.105 \\
zipf & $s=1.5$ & 1000 & 0.018 & 0.019 & 0.026 & \textbf{0.017} & 0.068 \\
\bottomrule
\end{tabular}
\caption{UCR (lower is better) for missing-mass estimation on Zipf populations across tail exponent $s$ and sample size $N$.}
\label{tab:synth_ucr_mass_zipf}
\end{table}

\subsection{Head--Tail mixtures}
\label{app:HT_dist}

Head--tail populations assign fraction $\rho$ of mass to a head of $S/2$ species and $1-\rho$ to a uniform tail. Larger $\rho$ concentrates probability on frequent types, which can stabilize singleton counts relative to Zipf/Dirichlet tails---the main synthetic regime where GT occasionally matches or beats our adjustment on missing mass. Table~\ref{tab:synth_ucr_mass_halfhalf} reports UCR for missing-mass estimation on Head--Tail populations.

\begin{table}[htbp]
\centering
\scriptsize
\setlength{\tabcolsep}{3.5pt}
\renewcommand{\arraystretch}{1.05}
\begin{tabular}{llrrrrrr}
\toprule
family & parameter & $N$ & Ours & GT & SGT & BD & PY \\
\midrule
HT & $\rho=0.7$ & 25 & 0.145 & \textbf{0.144} & \textbf{0.144} & 0.520 & 0.992 \\
HT & $\rho=0.7$ & 100 & 0.119 & \textbf{0.118} & 0.165 & 0.425 & 0.862 \\
HT & $\rho=0.7$ & 1000 & 0.033 & \textbf{0.032} & 0.311 & 0.098 & 0.149 \\
HT & $\rho=0.85$ & 25 & 0.154 & \textbf{0.153} & \textbf{0.153} & 0.512 & 0.977 \\
HT & $\rho=0.85$ & 100 & \textbf{0.117} & 0.119 & 0.228 & 0.378 & 0.804 \\
HT & $\rho=0.85$ & 1000 & \textbf{0.036} & \textbf{0.036} & 0.210 & 0.095 & 0.132 \\
HT & $\rho=0.95$ & 25 & \textbf{0.164} & \textbf{0.164} & 0.164 & 0.521 & 0.984 \\
HT & $\rho=0.95$ & 100 & \textbf{0.119} & \textbf{0.119} & 0.217 & 0.340 & 0.754 \\
HT & $\rho=0.95$ & 500 & 0.062 & \textbf{0.061} & 0.333 & 0.112 & 0.225 \\
HT & $\rho=0.95$ & 1000 & 0.022 & \textbf{0.021} & 0.112 & 0.102 & 0.075 \\
\bottomrule
\end{tabular}
\caption{Representative UCR values for missing-mass estimation on head--tail mixtures ($\rho$ = head-mass fraction). GT is within 0.001--0.002 of Ours at most settings; both dominate SGT and Bayesian baselines at small $N$.}
\label{tab:synth_ucr_mass_halfhalf}
\end{table}

\subsection{Normalized pairwise gain}

\begin{figure}[H]
    \centering

    \begin{minipage}[t]{0.48\textwidth}
        \centering
        \includegraphics[width=\linewidth]
        {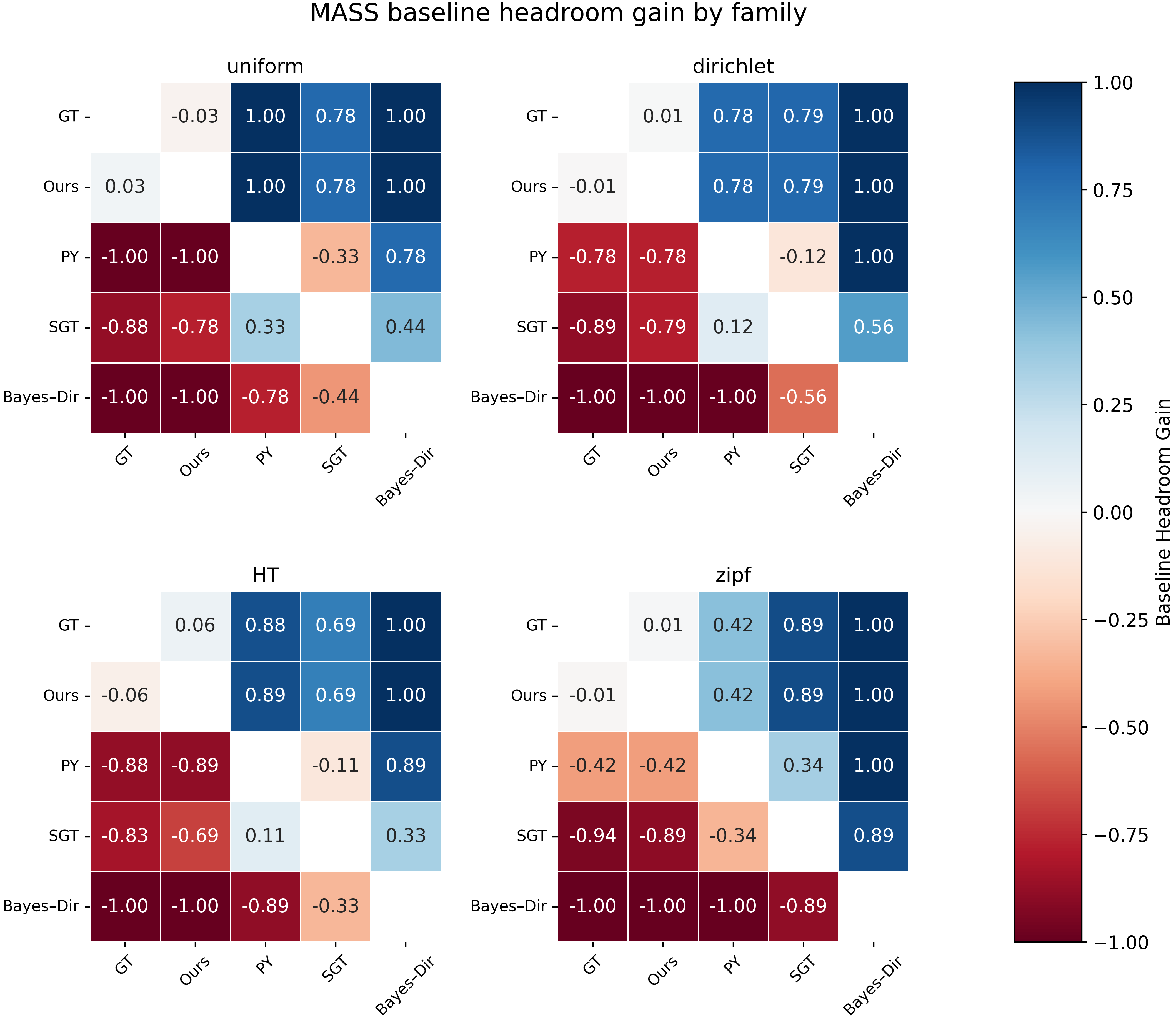}
        \par (a) Missing mass
    \end{minipage}
    \hfill
    \begin{minipage}[t]{0.48\textwidth}
        \centering
        \includegraphics[width=\linewidth]
        {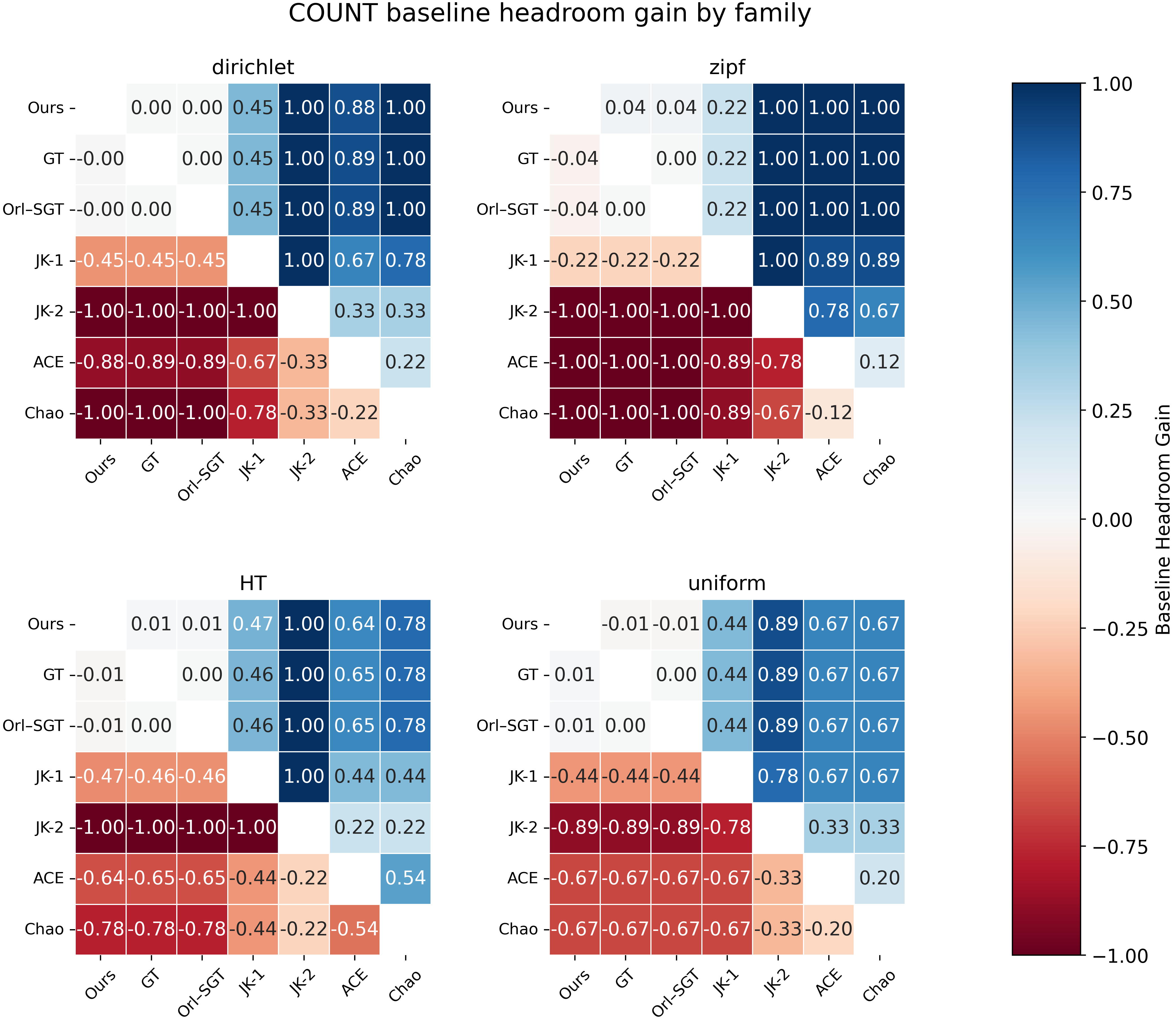}
        \par (b) Unseen-species prediction
    \end{minipage}

    \caption{Per-family normalized pairwise gain
    for (a) missing-mass and
    (b) unseen-species estimation.}
    \label{fig:synth_gain_family}
\end{figure}

Fig.~\ref{fig:synth_gain_family} presents the per-family
normalized pairwise gains for missing-mass and unseen-species
estimation. For missing-mass (Fig.~\ref{fig:synth_gain_family}a),
larger gains are observed in Zipf and Dirichlet distributions,
where rare-event statistics are more fragile, while head--tail
mixtures exhibit smaller improvements as GT remains competitive.
For unseen-species estimation (Fig.~\ref{fig:synth_gain_family}b),
the fragility-aware method achieves substantial gains over
classical estimators such as Chao, ACE, and jackknife methods,
highlighting its effectiveness in reducing estimation error
under fragmented, small-sample regimes.

\section{Extended Evaluation of Fragility-Aware Unseen-Species Estimation}
\label{App:extended_results}

\subsection{Shakespeare Corpus: Missing Mass Estimation}
\label{app:real_results_shake_mass}
\begin{figure}[H]
  \centering
  \includegraphics[width=\figwide]{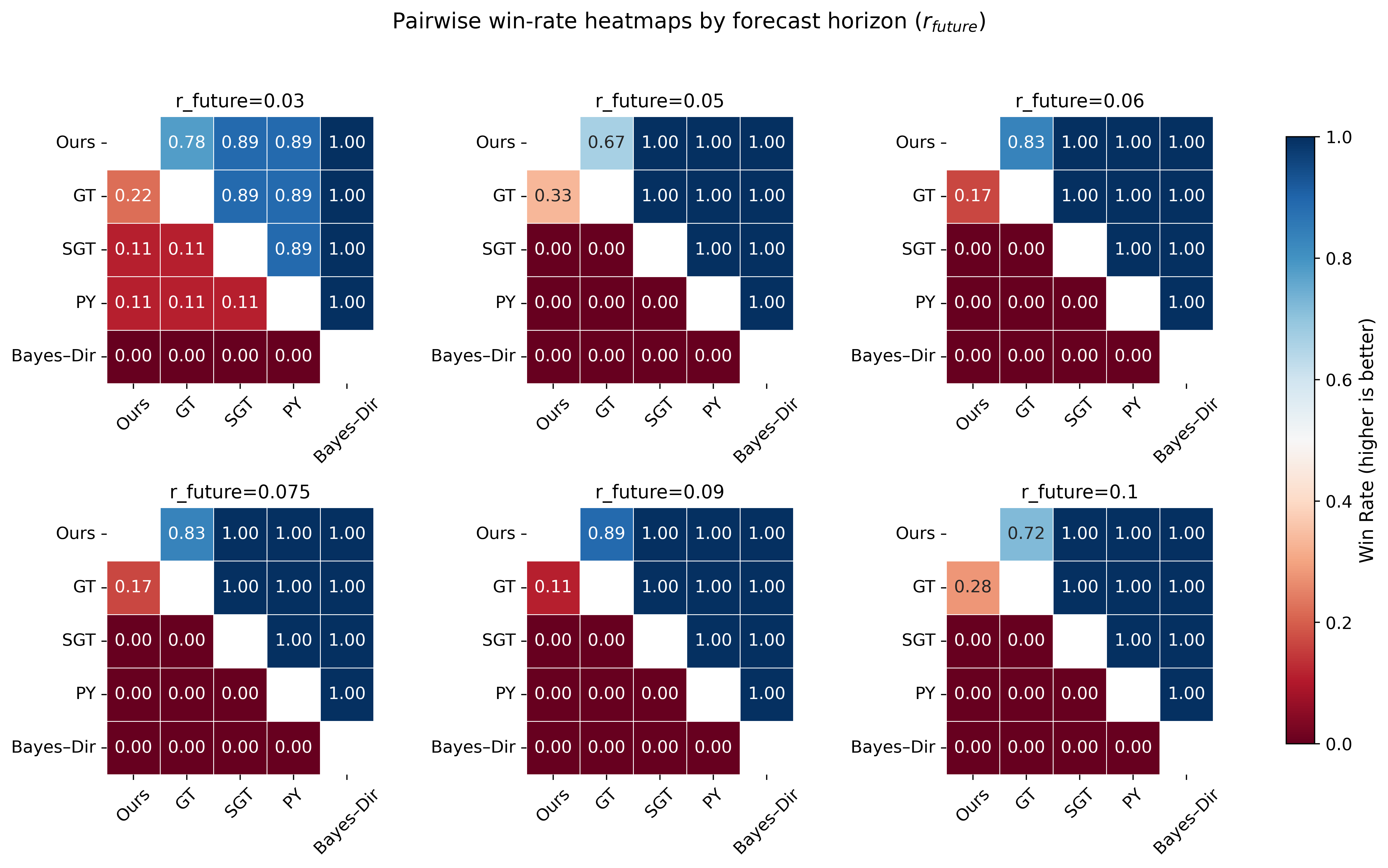}
  \vspace{-0.25em}
  \caption{Missing-mass win-rate heatmaps on Shakespeare across $r_{\mathrm{future}}\in[0.03,0.10]$ (fraction of experiments where the row beats the column on UCR).}
  \label{fig:shake_family_mass}
\end{figure}

Fig.~\ref{fig:shake_family_mass} shows missing-mass win rates across horizons $r_{\mathrm{future}}$. The fragility-aware estimator stays competitive with classical methods as the horizon grows, while singleton-driven estimators vary more.

\subsection{Shakespeare Corpus: Missing Count Estimation}
\label{app:real_results_shake_count}

\begin{figure}[H]
  \centering
  \includegraphics[width=\figwide]{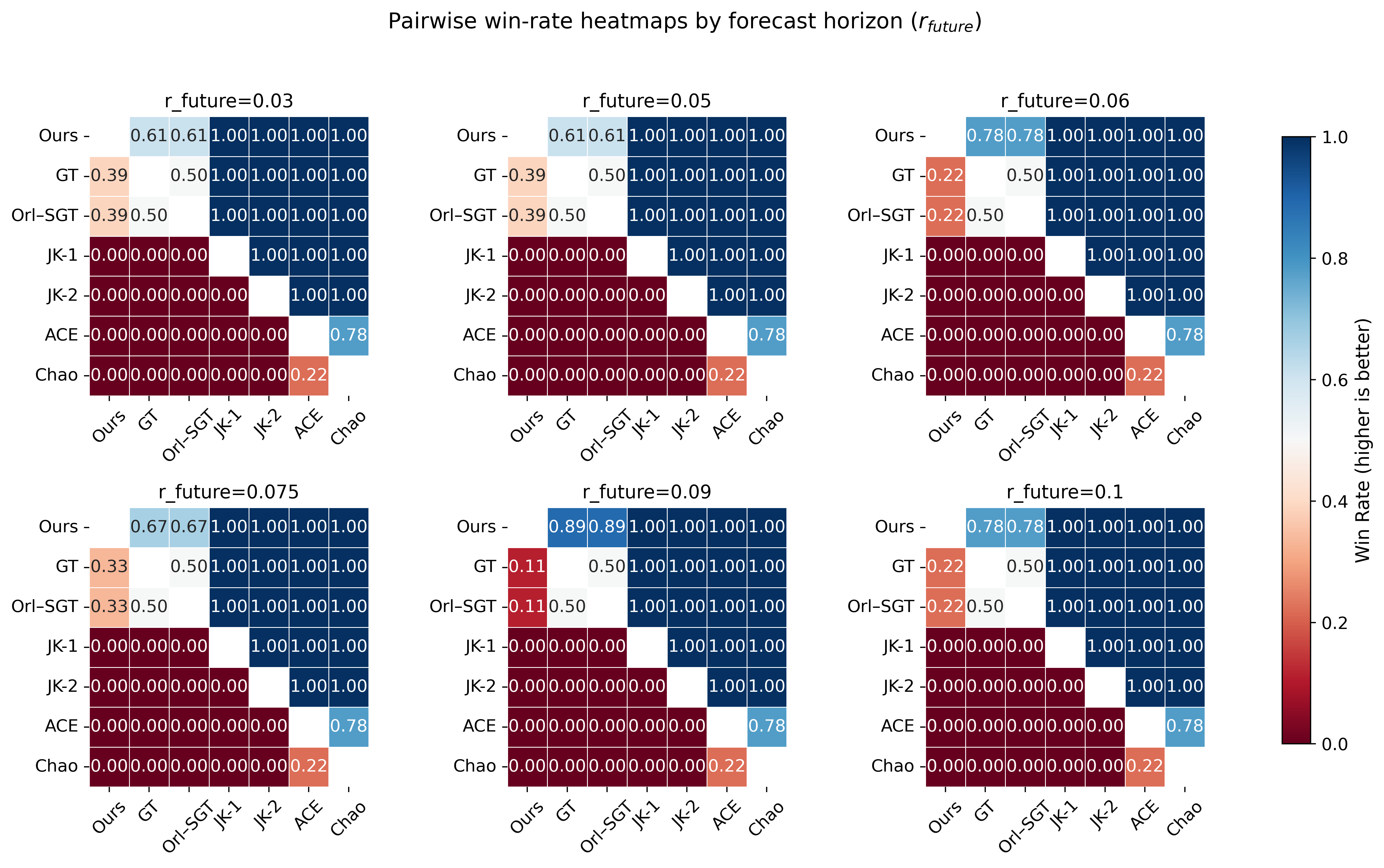}
  \caption{Unseen-count win-rate heatmaps on Shakespeare across $r_{\mathrm{future}}\in[0.03,0.10]$.}
  \label{fig:shake_family_count}
\end{figure}

Fig.~\ref{fig:shake_family_count} shows unseen-count win rates across horizons. The fragility-aware estimator remains robust under long-tailed extrapolation where rare-count baselines degrade.

\subsection{Google n-gram Corpus: Missing Mass Estimation}
\label{app:real_results_google_mass}

\begin{figure}[H]
  \centering
  \includegraphics[width=\figwide]{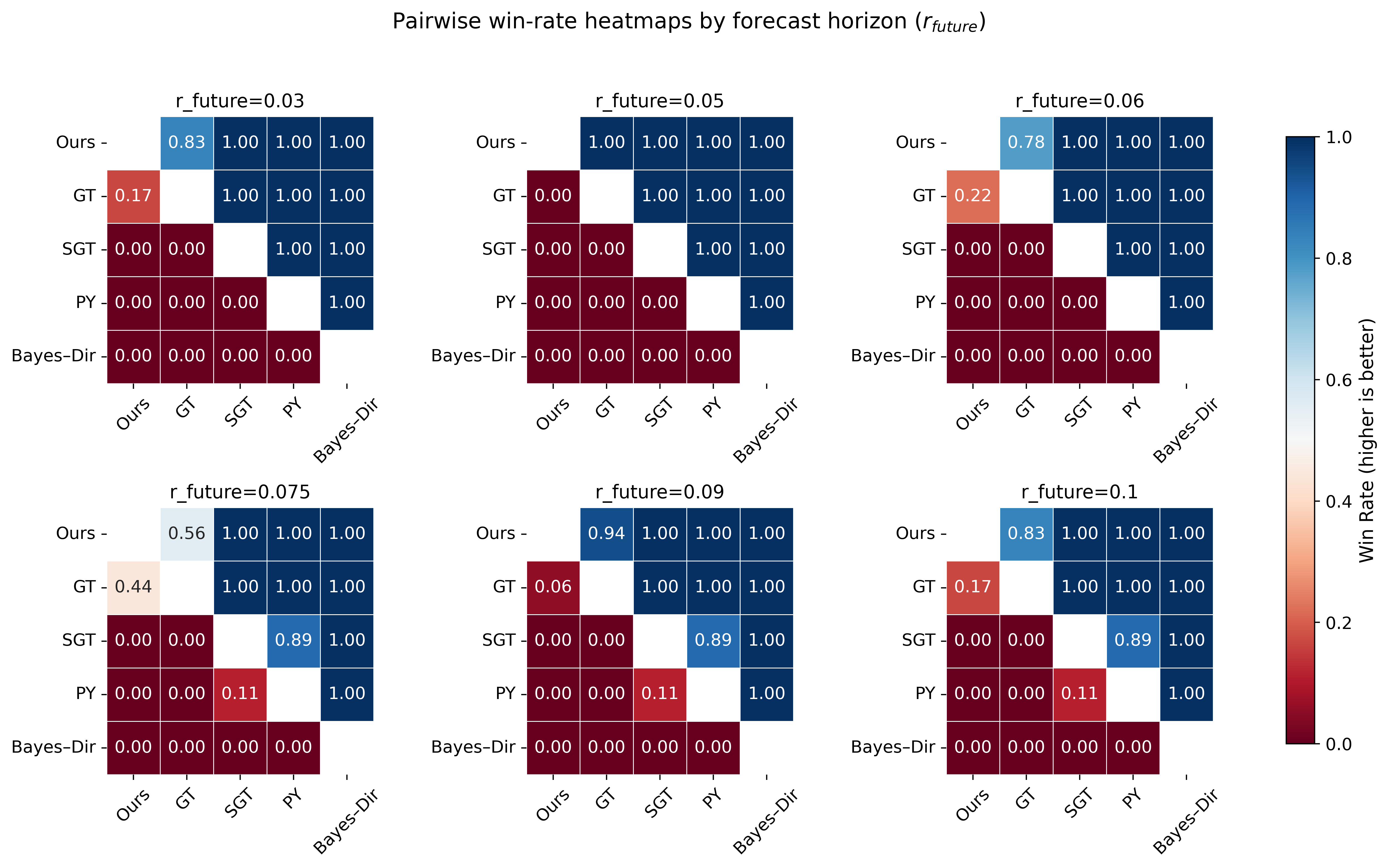}
  \vspace{-0.25em}
  \caption{Missing-mass win-rate heatmaps on Google 1-gram across $r_{\mathrm{future}}\in[0.03,0.10]$.}
  \label{fig:google_family_mass}
\end{figure}

Fig.~\ref{fig:google_family_mass} shows missing-mass win rates on Google 1-gram. The fragility-aware estimator attains the highest win rates across horizons, while frequency-driven estimators are more horizon-sensitive in this heavy-tailed corpus.

\subsection{Google n-gram Corpus: Missing Count Estimation}
\label{app:real_results_google_count}

\begin{figure}[H]
  \centering
  \includegraphics[width=\figwide]{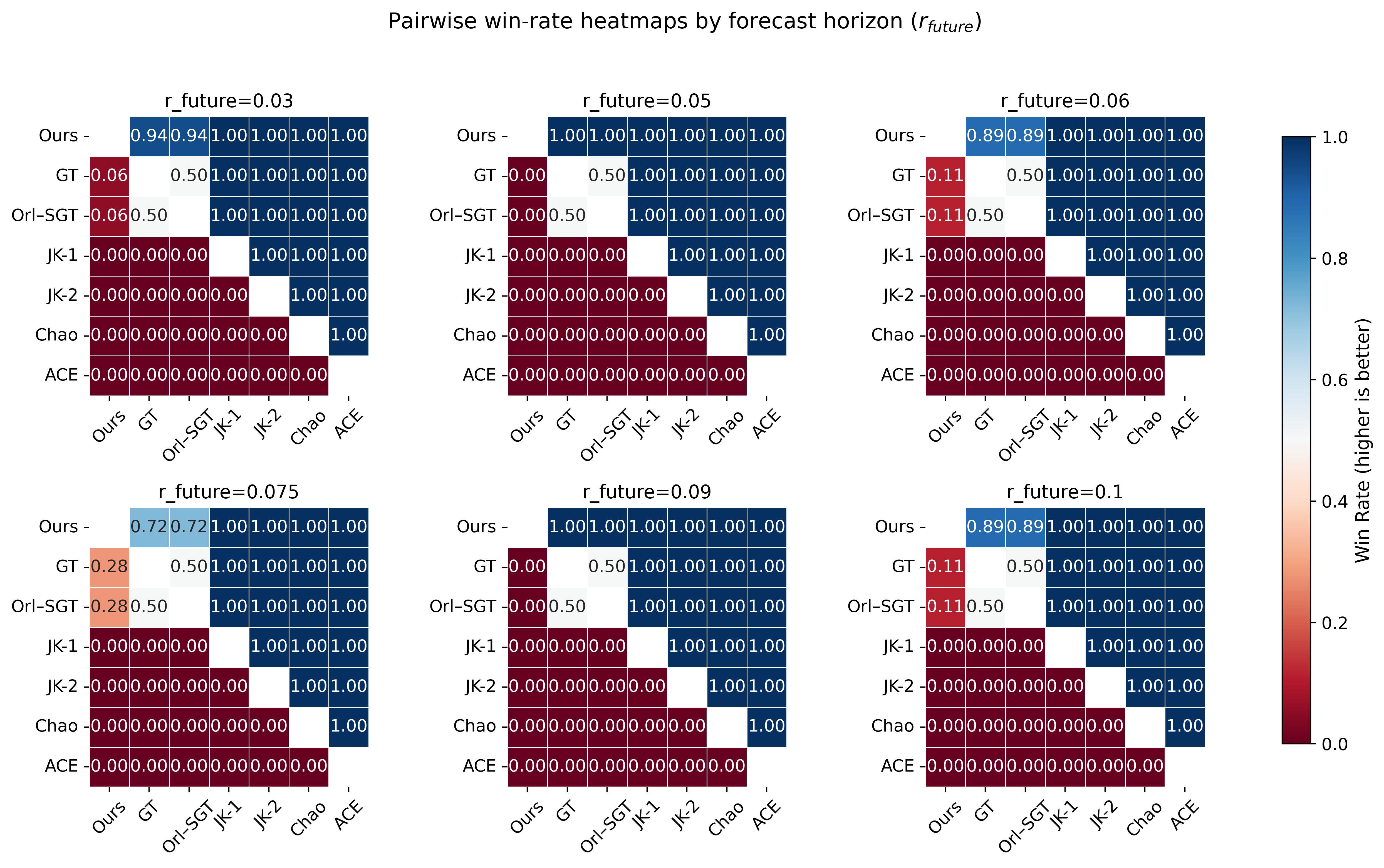}
  \caption{Unseen-count win-rate heatmaps on Google 1-gram across $r_{\mathrm{future}}\in[0.03,0.10]$.}
  \label{fig:google_family_count}
\end{figure}

Fig.~\ref{fig:google_family_count} shows unseen-count win rates on Google 1-gram. Explicit fragility modeling improves robustness as the extrapolation horizon grows.

\bibliographystyle{IEEEtran}
\bibliography{references} 